\documentclass[sigconf,nonacm]{acmart}

\renewcommand\footnotetextcopyrightpermission[1]{}
\setcopyright{none}

\AtBeginDocument{%
  }

\setcopyright{acmlicensed}
\copyrightyear{2026}
\acmYear{2026}
\acmDOI{XXXXXXX.XXXXXXX}  

\usepackage{xspace}
\newcommand{\ours}{\textsc{VeriSciQA}\xspace}

\usepackage{amsmath,amsfonts,bm}

\def\eqref#1{equation~\ref{#1}}

\def\1{\bm{1}}

\DeclareMathAlphabet{\mathsfit}{\encodingdefault}{\sfdefault}{m}{sl}
\SetMathAlphabet{\mathsfit}{bold}{\encodingdefault}{\sfdefault}{bx}{n}

\usepackage{enumitem}

\usepackage{subcaption}

\usepackage{tcolorbox}

\usepackage{pifont}

\usepackage{listings}

\usepackage{multirow}

\usepackage{placeins}

\newcommand{\appsecentry}[2]{%
  \par\noindent
  \hyperref[#2]{\ref*{#2}.\hspace{0.5em}#1}%
  \nobreak\dotfill\nobreak
  \hyperref[#2]{\makebox[1.5em][r]{\pageref*{#2}}}%
  \par
}
\newcommand{\appsubentry}[2]{%
  \par\noindent\hspace{1.5em}%
  \hyperref[#2]{\ref*{#2}\hspace{0.5em}#1}%
  \nobreak\dotfill\nobreak
  \hyperref[#2]{\makebox[1.5em][r]{\pageref*{#2}}}%
  \par
}

\title{VeriSciQA: An Auto-Verified Dataset for Scientific Visual Question Answering}

\author{Yuyi Li}
\affiliation{%
  \institution{Sun Yat-sen University}
  \city{Guangzhou}
  \country{China}}

\author{Daoyuan Chen}
\affiliation{%
  \institution{Alibaba Group}
  \city{Hangzhou}
  \country{China}}

\author{Zhen Wang}
\affiliation{%
  \institution{Sun Yat-sen University}
  \city{Guangzhou}
  \country{China}}

\author{Yutong Lu}
\affiliation{%
  \institution{Sun Yat-sen University}
  \city{Guangzhou}
  \country{China}}

\author{Yaliang Li}
\affiliation{%
  \institution{Alibaba Group}
  \city{Hangzhou}
  \country{China}}

\renewcommand{\shortauthors}{Li et al.}

\begin{abstract}
Large Vision-Language Models (LVLMs) show promise for scientific applications, yet open-source models still struggle with Scientific Visual Question Answering (SVQA), namely answering questions about figures from scientific papers. A key bottleneck is the lack of public, large-scale, high-quality SVQA datasets. Although recent work uses LVLMs to synthesize data at scale, we identify systematic errors in their resulting QA pairs, stemming from LVLMs' inherent limitations and information asymmetry between figures and text.
To address these challenges, we propose a Cross-Modal verification framework that generates questions and answers purely from figure-citing paragraphs, then verifies them against the figures themselves, leveraging the inherent text-figure alignment in scientific papers to filter out erroneous QA pairs.
We instantiate this framework to curate \ours, a dataset of 20{,}272 QA pairs spanning 20 scientific domains and 12 figure types. Difficulty assessment reveals a notable accuracy gap between the best open-source model (65\%) and the best proprietary model (80.5\%), demonstrating room for improvement. Moreover, models fine-tuned on \ours achieve consistent improvements on SVQA benchmarks, with performance gains that scale with data size, surpassing models trained on existing datasets. Human evaluation further validates the improved quality of \ours.
These results demonstrate that continued data expansion via our scalable framework can further advance SVQA capability in the open-source community.
Our dataset is publicly available at \url{https://huggingface.co/datasets/datajuicer/VeriSciQA}.
\end{abstract}

\begin{document}
\raggedbottom

\maketitle

\section{Introduction}\label{sec:intro}

Recent advances in Large Vision-Language Models (LVLMs)~\cite{alayrac2022flamingo,liu2023visual,bai2023qwenvlversatilevisionlanguagemodel,OpenAI_GPT4V_SystemCard_2023} have achieved strong performance on general-purpose vision-language tasks.
However, scientific-figure understanding remains a major bottleneck, as these figures encode dense quantitative evidence (e.g., trends, comparisons, ablations, and error bars) that requires specialized visual reasoning~\cite{wang2024charxiv}.
Reliable figure understanding is a prerequisite for building research assistants that can answer figure-grounded questions, cross-check reported results, and support reproducible scientific analysis.

To this end, we focus on the task of \emph{Scientific Visual Question Answering} (SVQA), where an LVLM answers questions grounded in figures from scientific papers.
While proprietary LVLMs are reported to approach human performance on SVQA benchmarks~\cite{wang2024charxiv}, open-source models still lag substantially behind their proprietary counterparts. A plausible cause is the lack of large-scale, high-precision SVQA training data that is both diverse and visually grounded.

Despite the growing interest in SVQA, existing dataset construction methods face a persistent trade-off among scalability, diversity, and correctness.
Template-based efforts such as FigureQA~\cite{kahou2017figureqa} and SciFiBench~\cite{roberts2024scifibench} achieve scalability but remain constrained by fixed question formats, limiting diversity.
Manually curated datasets like CharXiv~\citep{wang2024charxiv} ensure correctness through expert annotation, albeit at the cost of scalability and diversity.
Recent LVLM-based methods, including ArXivQA~\citep{li2024multimodal}, SPIQA~\citep{pramanick2024spiqa}, and LiveXiv~\citep{shabtay2024livexiv}, improve scalability and diversity, yet LVLM-driven generation usually introduces errors due to hallucination when interpreting visual inputs~\cite{bai2024hallucination}.
Despite this limitation, LVLM-based methods still offer the best prospect for scaling diverse SVQA data, but their correctness remains an open challenge.

Among LVLM-based methods, ArXivQA typifies this line of work, and its resulting SVQA dataset has been adopted in some follow-up studies~\cite{tong2024cambrian, dai2024nvlm, zhang2024beyond}. However, our systematic analysis of its SVQA dataset reveals four recurring categories of erroneous QA pairs (detailed in \S~\ref{sec:motivation}).
Collectively, these errors raise concerns about the reliability of LVLM-based methods, highlighting potential risks of fine-tuning LVLMs on their synthesized SVQA datasets.

We trace these errors to two primary sources.
(i)~\emph{Model-intrinsic hallucination}: LVLMs may fabricate visual evidence or trends that are not present in the figure, consistent with recent findings that even state-of-the-art LVLMs may hallucinate visual content~\cite{bai2024hallucination}.
(ii)~\emph{Information asymmetry}: the QA generation model only sees the figure (or figure+caption), while the figure-citing paragraphs define symbols, experimental settings, and the intended takeaway. This mismatch causes the LVLM to over-rely on its prior knowledge, producing questions whose answers are not entailed by the visual evidence.
These observations suggest two potential directions for improvement: (1) introducing verification mechanisms to detect and filter such hallucinations, and (2) incorporating figure-associated context to mitigate information asymmetry.

Several recent methods have made progress along both directions.
LiveXiv~\citep{shabtay2024livexiv} uses a second LVLM as a verifier to review each generated answer against the same figure, discarding QA pairs where the verifier disagrees with the generated answer.
SPIQA~\citep{pramanick2024spiqa} incorporates figure-citing paragraphs as additional context for QA generation.
However, same-figure LVLM agreement is an unreliable proxy for visual grounding: the generator and verifier, both operating on the same visual input, can share correlated hallucinations. Meanwhile, adding figure-citing paragraphs alone reduces information asymmetry but does not prevent visually unsupported QA pairs from entering the dataset.
To address both limitations, we decouple generation and verification across modalities: generate QA from figure-citing paragraphs to mitigate information asymmetry, and verify against figures to enforce visual grounding. This cross-modal design reduces correlated errors, since hallucinations from text-based generation and vision-based verification are less likely to coincide.

Concretely, we leverage two key properties of peer-reviewed papers: (1) figure-citing paragraphs typically reflect the authors' intended interpretation of the figure, and (2) many statements in these paragraphs, such as reported trends or numerical results, are verifiable against the figures.
Building on these properties, we propose a Cross-Modal Verification framework.
In this framework, figure-citing paragraphs are first fed into an LLM to produce questions and answers, mitigating the information asymmetry.
An LVLM then generates distractors conditioned on the figure to form multiple-choice QA pairs.
Auxiliary filters first target common error patterns, after which cross-modal verification discards QA pairs whose answers are not visually supported by the figure.

Our main contributions are as follows:
\begin{itemize}[leftmargin=*]
    \item \textbf{A Cross-Modal Verification framework for SVQA synthesis.} We propose a scalable framework that generates questions and answers from figure-citing paragraphs, produces plausible distractors via an LVLM to form multiple-choice QA pairs, and verifies visual grounding against figures, leveraging the inherent text-figure alignment in peer-reviewed scientific papers. 

    \item \textbf{A diverse, high-quality, and challenging SVQA dataset.} Built with our proposed framework, \ours comprises 20{,}272 QA pairs spanning 20 scientific domains, 12 figure types, and 5 question types (\S\ref{subsec:diversity}). Human evaluation confirms superior quality compared to existing SVQA datasets. Zero-shot evaluation on our dataset reveals a 15.9 percentage-point accuracy gap between the best proprietary and open-source models.

    \item \textbf{Scaling Data, Scaling Capability.} Models fine-tuned on our dataset achieve consistent improvements across diverse SVQA benchmarks, outperforming those trained on existing SVQA datasets. As our training data scales from 500 to 20{,}272 examples, fine-tuned models improve monotonically, reaching an average gain of +2.05\% over the base model.    
\end{itemize}
We release both the dataset and Data-Juicer \cite{dj2} operators that implement our framework (Appendix~\ref{app:data-pipeline}), facilitating the community to reproduce, reuse, and extend the proposed pipeline.

\section{Motivation}
\label{sec:motivation}

A prominent line of work in SVQA dataset construction relies on LVLM-based methods that prompt an LVLM with a figure from the paper, its accompanying caption, and optional context parsed from the paper’s LaTeX source. ArXivQA~\citep{li2024multimodal}, a representative work in this line, synthesizes more than 100\,K multiple-choice QA pairs. Despite its large scale, our audit reveals that it still exhibits a 37\% error rate, raising concerns about the data quality of SVQA datasets produced by such LVLM-based methods.

\begin{figure*}[!thbp]
  \centering
  \begin{subfigure}[t]{0.48\textwidth}
      \centering
      \includegraphics[height=4.0cm,keepaspectratio]{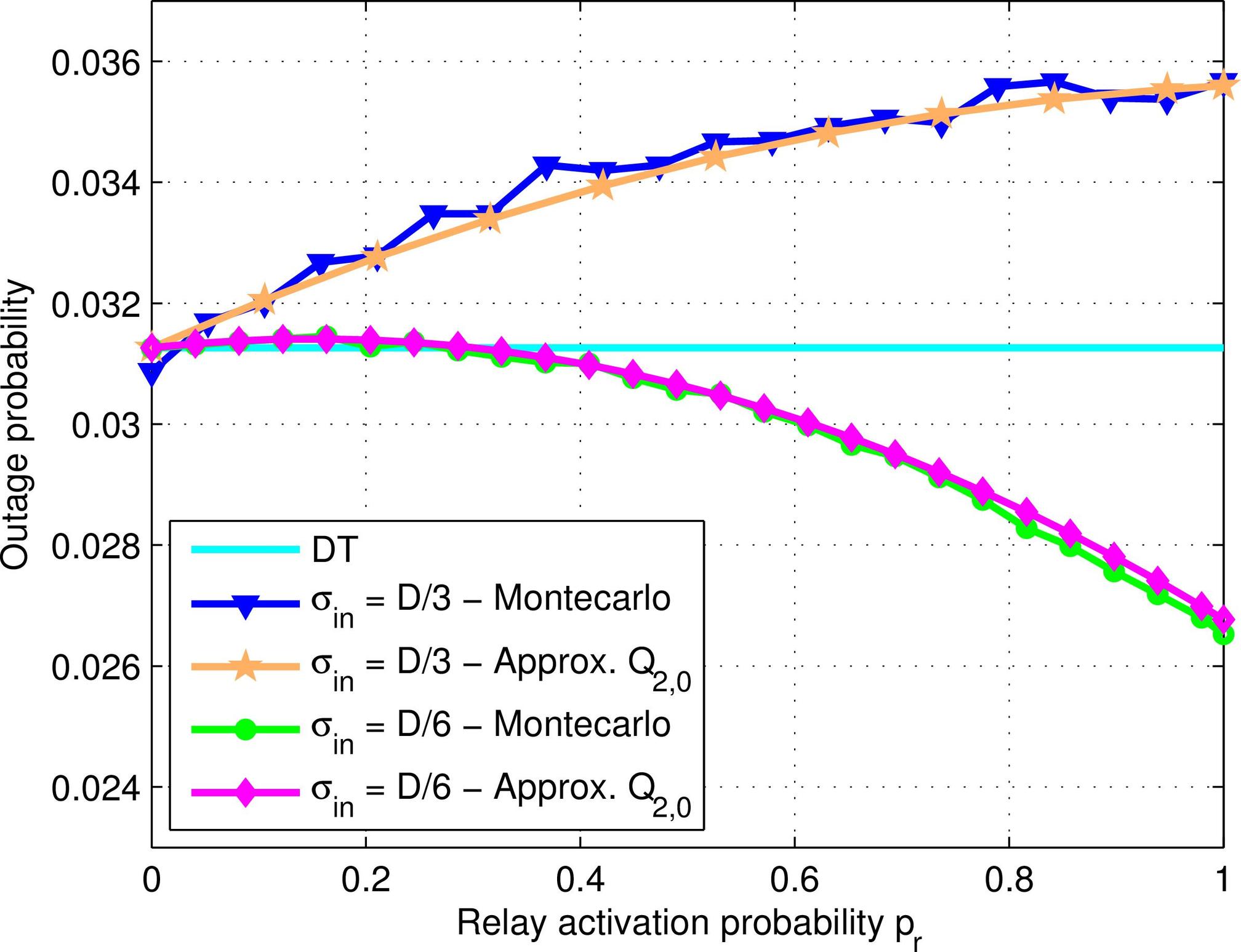}
      
      \vspace{3pt}
      \raggedright
      \footnotesize
      \textbf{Question:} At what relay activation probability does the \(\sigma_{in} = D/6\) scenario reach lowest outage?\\
      \textbf{Options:} A. 0.2, B. 0.4, C. 0.6, D. 1.0 \\
      \textbf{Label:} C. 0.6 \hfill \\
      \textbf{Rationale:} The green line shows outage probability decreasing as relay activation increases, reaching the lowest point at around 0.6 before plateauing.
      \caption{E1: Incorrectly visually grounded}
      \label{fig:arxivqa-e1}
  \end{subfigure}
  \hfill
  \begin{subfigure}[t]{0.48\textwidth}
      \centering
      \includegraphics[height=4.0cm,keepaspectratio]{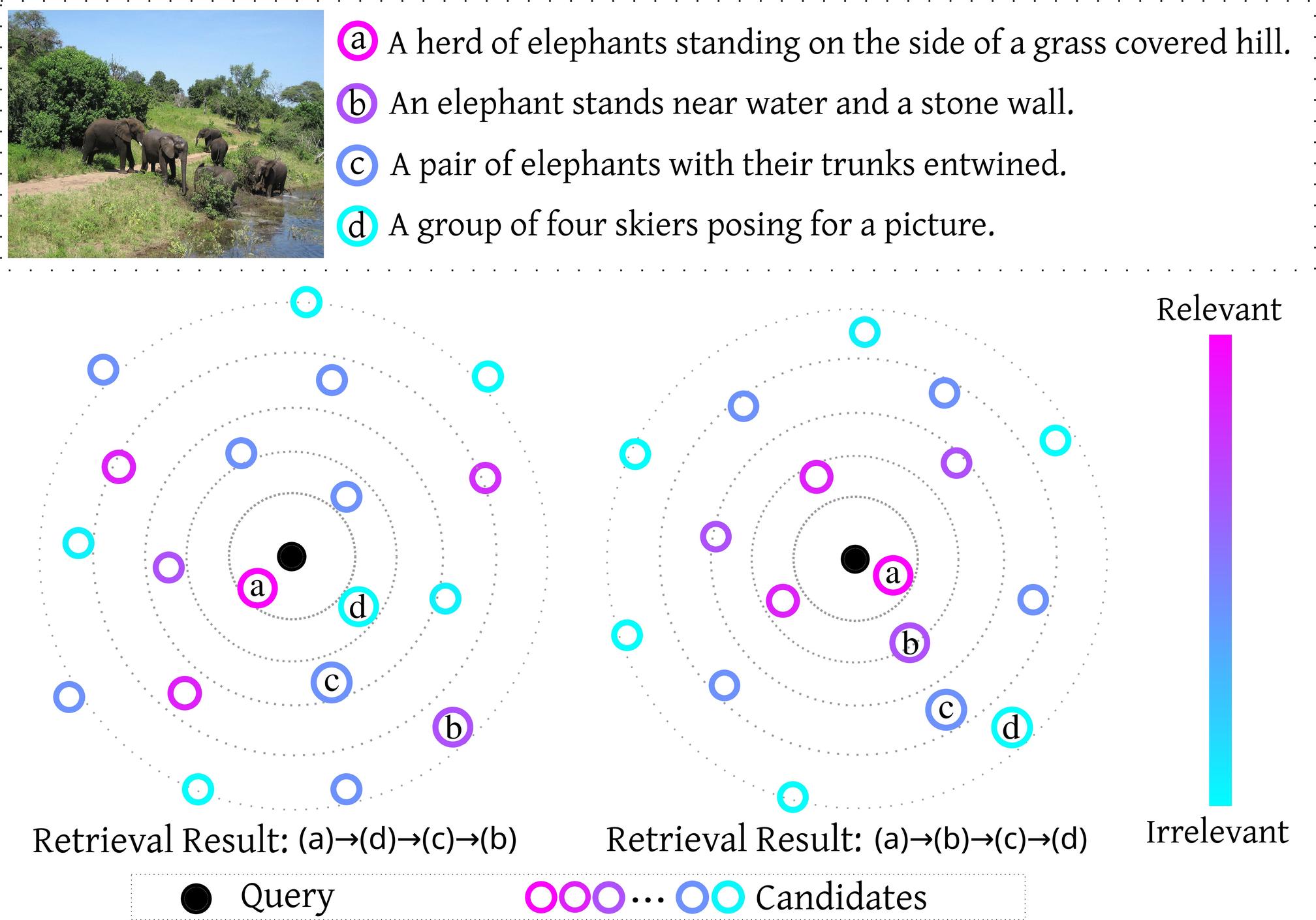}
      
      \vspace{3pt}
      \raggedright
      \footnotesize
      \textbf{Question:} What is the primary subject of the image?\\
      \textbf{Options:} A. Elephants on a hill, B. Elephant near water and wall, C. Elephants with trunks entwined, D. Four skiers posing \\
      \textbf{Label:} B \hfill \\
      \textbf{Rationale:} The image shows a single elephant near water with a stone structure. Choices A and C are incorrect because they imply multiple elephants. Choice D is irrelevant as it doesn't feature skiers.
  \caption{E2: Figure-intent misaligned}
  \label{fig:arxivqa-e2}
  \end{subfigure}
  
  \vspace{10pt}
  
  \begin{subfigure}[t]{0.58\textwidth}
      \centering
      \includegraphics[height=4.8cm,keepaspectratio]{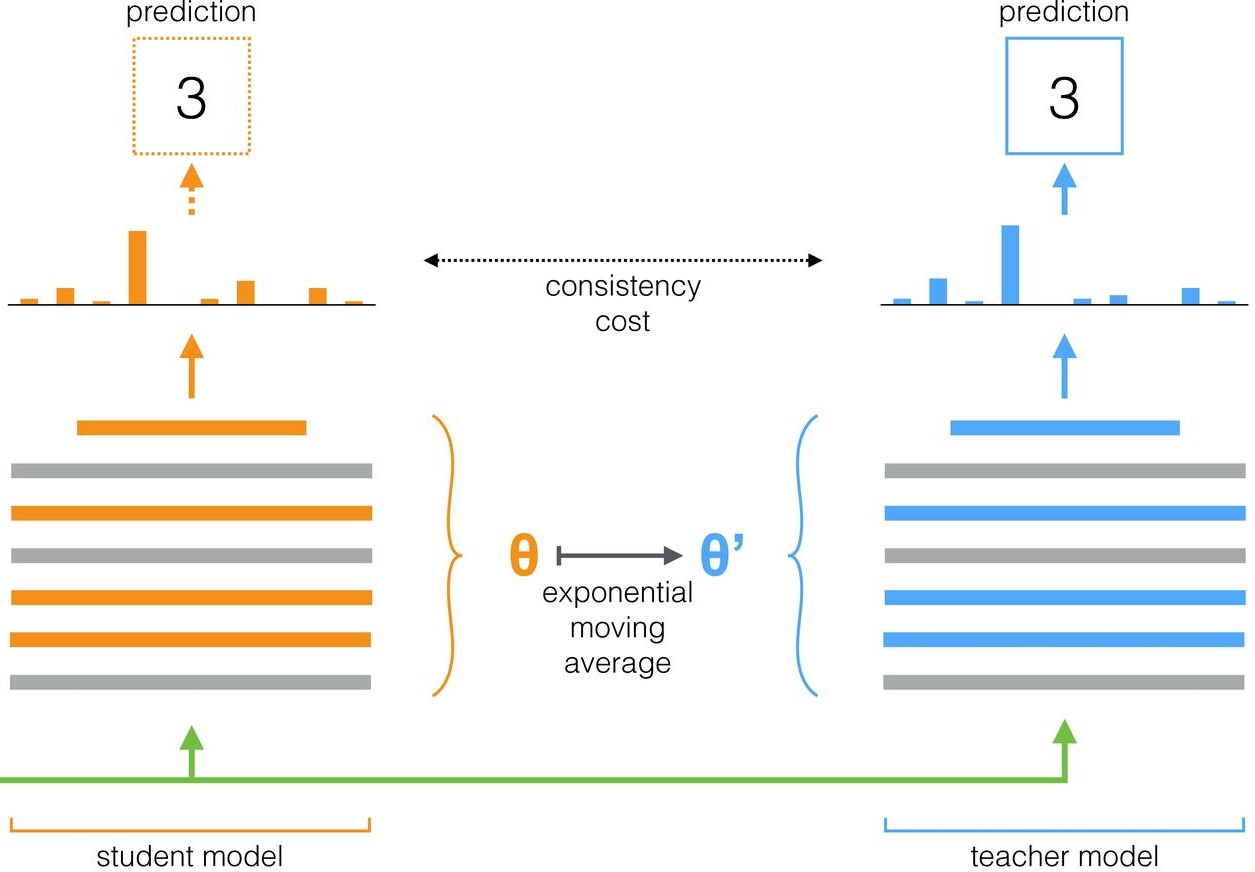}
      
      \vspace{3pt}
      \raggedright
      \footnotesize
      \textbf{Question:} What does "exponential moving average" refer to?\\
      \textbf{Options:} A. Smoothing technique for trends, B. Calculating median values, C. Forecasting stock prices, D. Averaging chemical mixtures \\
      \textbf{Label:} A \hfill \\
      \textbf{Rationale:} The 'exponential moving average' is a commonly used technique in time series analysis to smooth out data.
      \caption{E3: Non-visual}
      \label{fig:arxivqa-e3}
  \end{subfigure}
  \hfill
  \begin{subfigure}[t]{0.38\textwidth}
      \centering
      \includegraphics[height=4.8cm,keepaspectratio]{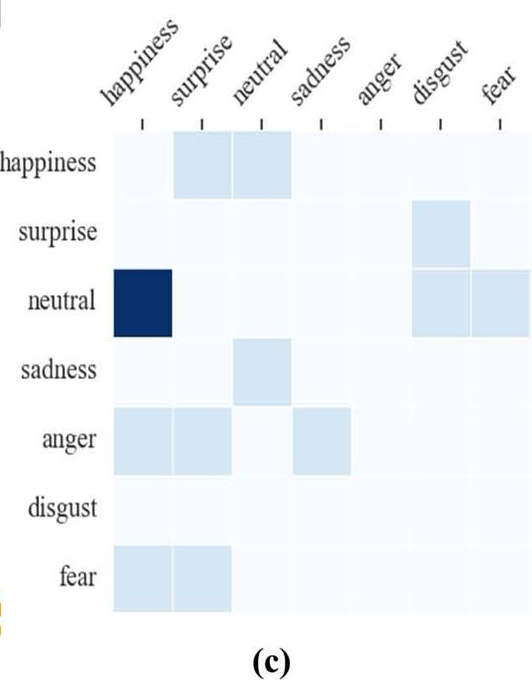}
      
      \vspace{3pt}
      \raggedright
      \footnotesize
      \textbf{Question:} Which emotion pair is least likely to co-occur?\\
      \textbf{Options:} A. Happiness-surprise, B. Anger-disgust, C. Sadness-neutral, D. Disgust-fear \\
      \textbf{Label:} D \hfill \\
      \textbf{Rationale:} The lightest square is between disgust and fear, suggesting these emotions are least likely to co-occur.
      \caption{E4: Outside-knowledge}
      \label{fig:arxivqa-e4}
  \end{subfigure}
  
  \caption{Illustrative examples of the four common error categories (E1–E4).}
  \Description{Four subfigures showing error examples from ArXivQA dataset: (a) incorrectly visually grounded error with a relay activation probability chart, (b) figure-intent misaligned error with an elephant image misinterpreted, (c) non-visual error where the answer can be inferred without the figure, and (d) outside-knowledge error requiring paper context to interpret a confusion matrix.}
  \label{fig:arxivqa-errors}
  \end{figure*}

\noindent\textbf{Systematic error analysis of ArXivQA.}
We randomly sample 100 items from ArXivQA, each released as a five‑tuple $\langle$\texttt{Question, Options, Label, Rationale, Figure}$\rangle$, where \texttt{Label} refers to the LVLM’s prediction rather than ground truth. Our systematic analysis reveals four common categories of erroneous QA pairs, which we refer to as follows:
\begin{itemize}[leftmargin=*]\setlength\itemsep{2pt}
\item \textbf{(E1) Incorrectly Visually Grounded} (19/100):
      questions or answers that exhibit mismatches between textual claims and the visual evidence~\citep{bai2024hallucination}, e.g., misreading chart values (Fig.~\ref{fig:arxivqa-e1}).
    \item \textbf{(E2) Figure–Intent Misaligned} (7/100): questions whose semantics diverge from the figure's purpose, e.g., treating a figure illustrating an embedding space as a natural image and asking for scene classification (Fig.~\ref{fig:arxivqa-e2}).
    \item \textbf{(E3) Non-Visual} (6/100):
          answers can be inferred directly from the question and common/domain-specific knowledge, making the figure unnecessary (Fig.~\ref{fig:arxivqa-e3}).
    \item \textbf{(E4) Outside-Knowledge} (5/100):
          questions that reference visual elements whose interpretation requires additional context from the paper. Since this context is not included in the question, the question cannot be reliably answered from the figure alone (Fig.~\ref{fig:arxivqa-e4}).
\end{itemize}

\noindent\textbf{Root causes: hallucination and information asymmetry.}
LVLMs are known to \emph{hallucinate}, frequently producing responses inconsistent with the visual content provided (i.e., cross-modal inconsistency)~\cite{bai2024hallucination}. 
The first two failure modes, E1 and E2, stem directly from this phenomenon. Methods such as ArXivQA rely solely on LVLM generation without verification, allowing hallucinated responses to pass through.
While LiveXiv~\citep{shabtay2024livexiv} uses a second LVLM to verify generated answers, both the generator and the verifier operate on the same visual input and may share similar hallucination patterns.
The other root cause is \emph{information asymmetry}. The input to the LVLM often contains only the figure and its caption, omitting essential figure-associated context that authors describe in the main body of the paper. 
Without this context, the model may over-rely on its broad knowledge and generate QA pairs that extend beyond what the figure alone conveys, leading to E2–E4 errors. 
SPIQA~\citep{pramanick2024spiqa} incorporates figure-citing paragraphs as additional context for QA generation, but without reliable verification, erroneous QA pairs may persist.
Thus, SPIQA and LiveXiv each target one root cause but not the other, motivating a unified approach that combines enriched context with verification, enabled by figure-citing paragraphs.

\noindent\textbf{Key insight: figure-citing paragraphs contain verifiable claims about figures.} 
Figure-citing paragraphs, which can be extracted from the paper's LaTeX source by matching figure reference commands (e.g., \texttt{\textbackslash cref}), often contain verifiable claims such as ``As shown in Fig.~2, accuracy rises by 20\%.''
In this example, the claim suggests that the figure likely aims to report model accuracy, and asserts a specific trend that can be checked against the figure.
More generally, claims in figure-citing paragraphs exhibit two key properties: they tend to reflect the authors' intended interpretation of the figure, and they often describe observable properties (e.g., trends or numerical results) that are verifiable against the figure.
These properties give rise to \emph{Cross-Modal Verification}: questions and answers are first generated from such claims, promoting alignment with the authors' intent; the generated answers can then be verified against the figure, helping ensure visual grounding. Since hallucinations from text-based generation and those from vision-based verification are less likely to be correlated, this cross-modal framework, together with auxiliary filters, helps systematically mitigate both hallucination and information asymmetry errors (E1--E4).

\begin{figure*}[t]
\centering
\includegraphics[width=\textwidth]{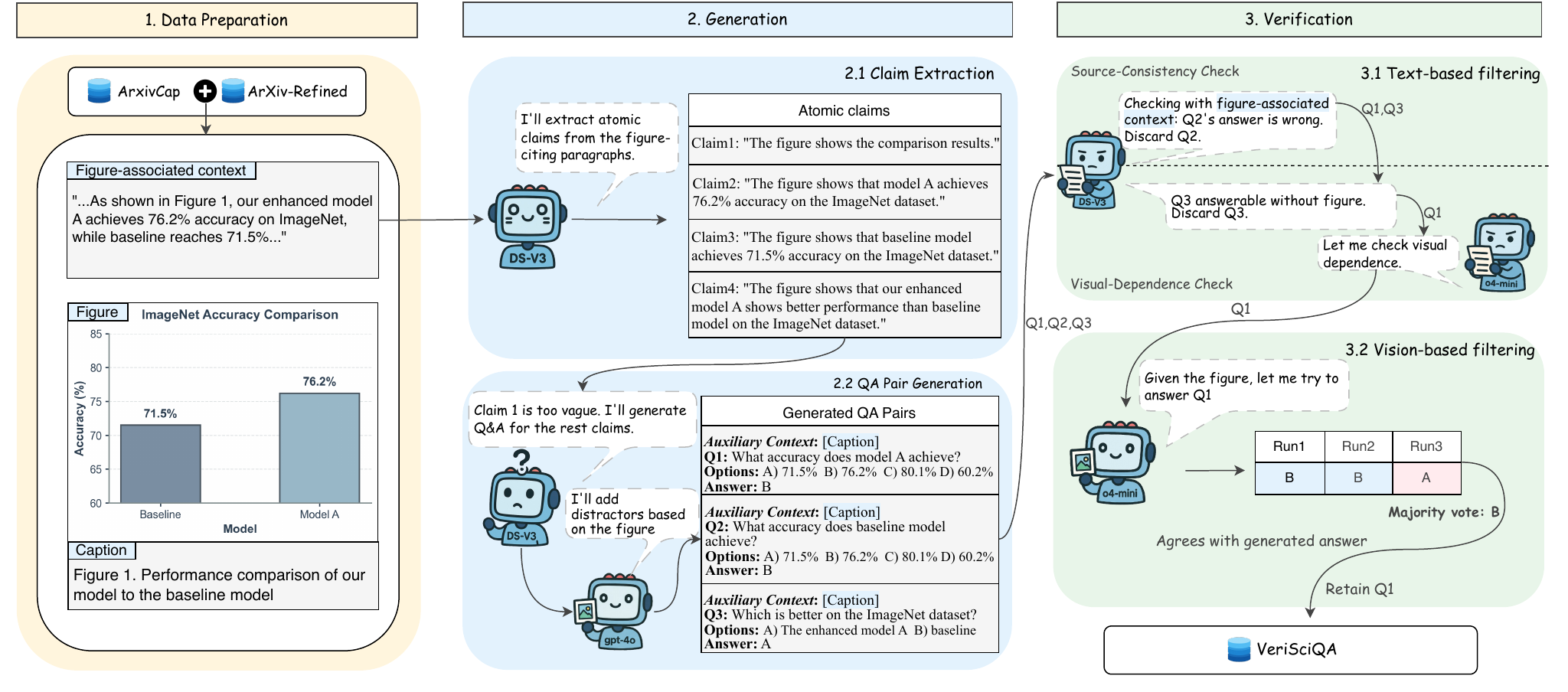}
\caption{Overview of our Cross-Modal Verification framework for curating \ours. We first extract figure-citing paragraphs from arXiv papers as figure-associated context. \textbf{Generation} stage decomposes each paragraph into atomic claims, generates questions and correct answers via an LLM, and produces visually grounded distractors via an LVLM to form multiple-choice QA pairs; \textbf{Verification} stage applies text-based filtering followed by vision-based filtering to remove erroneous QA pairs.}
\Description{A pipeline diagram showing the Cross-Modal Verification framework. The left side shows data preparation extracting figure-citing paragraphs. The middle shows the Generation stage with claim extraction, question-answer generation via an LLM, and distractor generation via an LVLM. The right side shows the Verification stage with cascaded filtering including source-consistency check, visual-dependence check, and visual-consistency check.}
\label{fig:method}
\end{figure*}

\section{Methodology}\label{sec:method}
To mitigate the four error types (E1--E4) identified in \S~\ref{sec:motivation}, we propose a \emph{Cross-Modal Verification} framework that generates questions and answers from figure-citing paragraphs, produces distractors conditioned on the figure, and then applies auxiliary filters and cross-modal verification to filter out erroneous QA pairs.

Fig.~\ref{fig:method} illustrates our framework, which comprises two stages: \textbf{Generation} and \textbf{Verification}, as instantiated to curate \ours from arXiv papers.
Starting from figure-citing paragraphs as figure-associated context, the \textbf{Generation} stage (\S~\ref{subsec:generation}) decomposes each paragraph into atomic claims and feeds them into a text-only LLM to generate a question and its correct answer from each claim, where the LLM skips claims that lack concrete visual grounding; an LVLM then generates plausible distractors conditioned on the figure to complete the multiple-choice format.
The \textbf{Verification} stage (\S~\ref{subsec:verification}) applies cascaded filtering: text-based filtering first checks for source-consistency and visual-dependence without accessing the figure; vision-based filtering then checks each remaining QA candidate against the figure using self-consistency~\citep{wang2022self} with an LVLM.
Only QA pairs that pass all checks are retained in the final dataset.
We detail each stage below.
\subsection{Framework Formalization}\label{subsec:formalization}

We formalize our framework to clarify the operational flow.
Let a scientific paper be represented as \(\mathcal{D} = (\mathcal{T}, \mathcal{F})\), where \(\mathcal{T}\) is the full text content and \(\mathcal{F} = \{(F_i, C_i)\}_{i=1}^N\) is the set of figure-caption pairs.
For each figure \(F_i\), we denote its corresponding \textit{figure-associated context} by \(\mathcal{P}_i \subset \mathcal{T}\), which consists of paragraph(s) in the document body that explicitly refer to the figure \(F_i\).
We process each figure identically and thus omit the index $i$ for brevity.

Let \(M_{\text{text}}\) denote an LLM, and let \(M_{\text{vision}}^{\text{gen}}\) and \(M_{\text{vision}}^{\text{verify}}\) denote two LVLMs capable of processing both images and text, one for distractor generation and one for verification, respectively.
In the \textit{Generation} stage, \(M_{\text{text}}\) first extracts a set of atomic claims \(\mathcal{S}=\{s_j\}\) from context \(\mathcal{P}\), then produces a question \(Q_j\) and correct answer \(A^*_j\) from each claim $s_j$.
Subsequently, \(M_{\text{vision}}^{\text{gen}}\) generates plausible distractor options based on \((Q_j, A^*_j, F, C)\), yielding the complete candidate QA pair \(q_j = (Q_j, O_j, A^*_j)\), where \(O_j\) is the set of all options including \(A^*_j\).
In the \textit{Verification} stage, we apply cascaded filtering with three boolean filters:
\begin{align*}
V_{\text{src}}&: (Q, O, A^*, \mathcal{P}) \xrightarrow{M_{\text{text}}} \{\text{True}, \text{False}\} \\
V_{\text{vis\_dep}}&: (Q, O, A^*, C) \xrightarrow{M_{\text{vision}}^{\text{verify}}} \{\text{True}, \text{False}\} \\
V_{\text{vis\_con}}&: (Q, O, A^*, F, C) \xrightarrow{M_{\text{vision}}^{\text{verify}}} \{\text{True}, \text{False}\}
\end{align*}
A candidate \(q_j\) is retained in the final dataset if and only if:
\begin{equation}
\begin{aligned}
&V_{\text{src}}(Q_j, O_j, A^*_j, \mathcal{P}) \land V_{\text{vis\_dep}}(Q_j, O_j, A^*_j, C) \\
&\quad\land V_{\text{vis\_con}}(Q_j, O_j, A^*_j, F, C) = \text{True}.
\end{aligned}
\label{eq:passcriterion}
\end{equation}

\subsection{Generation}\label{subsec:generation}

This stage consists of the following three steps:

\noindent\textbf{Claim Extraction.}
We instruct \(M_{\text{text}}\) to extract and reformulate factual statements about the figure $F$ from \(\mathcal{P}\) as atomic claims:
\begin{equation}
\mathcal{S} = \{s_j\} = M_{\text{text}}(\mathcal{P}),
\label{eq:claim-extraction}
\end{equation}
where each \(s_j\) denotes an atomic claim, a concise, self-contained sentence that asserts what \(F\) shows, following the pattern ``The figure shows...'' (see the illustrative examples in Fig.~\ref{fig:method}).

\noindent\textbf{Question \& Answer Generation.}
For each atomic claim \(s_j \in \mathcal{S}\), we instruct \(M_{\text{text}}\) to generate a question and its correct answer:
\begin{equation}
(Q_j, A^*_j) = M_{\text{text}}(s_j, C, \mathcal{P}),
\label{eq:qa-generation}
\end{equation}
where the model converts \(s_j\) into a question \(Q_j\) and derives the correct answer \(A^*_j\) from the claim.
Because each \(s_j\) explicitly describes what \(F\) shows, generating questions from it promotes figure-intent alignment, mitigating \emph{figure-intent misalignment} errors (E2).
\(M_{\text{text}}\) may decline to produce a question and answer if \(s_j\) is too vague or lacks sufficient grounding, e.g., Claim~1 in Fig.~\ref{fig:method}.

\noindent\textbf{Distractor Generation.}
To complete the multiple-choice format, we instruct \(M_{\text{vision}}^{\text{gen}}\) to generate three plausible distractor options:
\begin{equation}
O_j = M_{\text{vision}}^{\text{gen}}(Q_j, A^*_j, F, C),
\label{eq:distractor-generation}
\end{equation}
where \(O_j\) includes the correct answer \(A^*_j\) along with distractors.
By conditioning on both the figure \(F\) and the correct answer \(A^*_j\), the LVLM can produce distractors that are visually plausible yet incorrect.
Since captions typically contain the information needed to comprehend a figure, each QA pair is accompanied by its corresponding figure's caption \(C\), helping make the pair self-contained and mitigating \emph{outside-knowledge} errors (E4).

\subsection{Verification}\label{subsec:verification}

Each candidate \(q_j\) is retained only if it passes all three filters, as defined in Eq.~\ref{eq:passcriterion}.
We detail the implementation of each filter below.

\noindent\textbf{Text-Based Filtering.}
The first step applies two text-only filters without accessing figure \(F\):

\begin{enumerate}[leftmargin=1.4em,itemsep=2pt]
  \item \textbf{Source-Consistency Check} \(V_{\text{src}}(Q_j, O_j, A^*_j, \mathcal{P})\): We prompt \(M_{\text{text}}\) to select an answer from \(O_j\) given context \(\mathcal{P}\) and question \(Q_j\). The filter returns \textbf{True} if the model uniquely identifies \(A^*_j\), and \textbf{False} otherwise, i.e., when it fails to answer, selects multiple options, or chooses a different answer (see Q2 in Fig.~\ref{fig:method}). This helps ensure the generated QA pair maintains consistency with the source context.
  
  \item \textbf{Visual-Dependence Check} \(V_{\text{vis\_dep}}(Q_j, O_j, A^*_j, C)\): We prompt \(M_{\text{vision}}^{\text{verify}}\) to select an answer from \(O_j\) given only caption \(C\) and question \(Q_j\) (without figure \(F\)). The filter returns \textbf{True} if the model fails to identify \(A^*_j\), and \textbf{False} otherwise (see Q3 dropped by visual-dependence check in Fig.~\ref{fig:method}). This helps ensure that correctly answering this question requires visual information from the figure, thereby mitigating \emph{non-visual} questions (E3).
\end{enumerate}

\noindent\textbf{Vision-Based Filtering.}
The final step verifies that \(A^*_j\) is visually grounded in \(F\) and that \(F\) and \(C\) together provide sufficient information to correctly answer \(Q_j\).
Since text-based filtering does not access figure \(F\), it cannot verify two critical aspects: (i) whether \(A^*_j\) is actually supported by visual evidence in \(F\), and (ii) whether the information in \(F\) and \(C\) is sufficient to answer \(Q_j\) without requiring additional context from the paper.
Therefore, we design \(V_{\text{vis\_con}}(Q_j, O_j, A^*_j, F, C)\) so as to prompt \(M_{\text{vision}}^{\text{verify}}\) to select an answer from \(O_j\) based solely on \(F\), \(C\), and \(Q_j\). The filter returns \textbf{True} if the model's prediction matches \(A^*_j\), and \textbf{False} otherwise (see Q1 accepted by vision-based filtering in Fig.~\ref{fig:method}). This helps mitigate \emph{incorrectly visually grounded} (E1) and \emph{outside-knowledge} (E4) errors.
Despite retaining only those \(q_j\) that \(M_{\text{vision}}^{\text{verify}}\) answers correctly, the resulting dataset remains challenging (\S\ref{subsec:difficulty}).
Importantly, while this step applies stringent filtering criteria at the cost of dataset quantity, our ablation study (\S\ref{subsec:ablation}) confirms that it is important for maintaining dataset quality.

\noindent\textbf{Mitigating Generation Bias.}
Our framework constrains generation bias at two levels.
First, each question \(Q_j\) and correct answer \(A^*_j\) are derived from an atomic claim \(s_j\) that \(M_{\text{text}}\) extracts from human-authored figure-citing paragraphs (Eq.~\ref{eq:claim-extraction}--\ref{eq:qa-generation}).
As a result, the topical coverage and question types of the resulting QA pairs are largely shaped by the source text rather than by model preferences, which constrains the bias of \(M_{\text{text}}\) during QA generation.
Second, we mitigate non-visual shortcuts in the options.
The visual-dependence filter \(V_{\text{vis\_dep}}\) removes QA pairs where language patterns in the distractors make \(A^*_j\) identifiable without inspecting the figure.
We also randomly shuffle option positions so that the correct answer is uniformly distributed across A/B/C/D, eliminating positional bias.

\subsection{Instantiation}\label{subsec:instantiation}

We instantiate our framework using scientific figures from peer-reviewed papers in arXiv, leveraging existing datasets and language models to implement each stage.

\noindent\textbf{Data Sources.}
We combine two complementary data sources:
ArXivCap~\citep{li2024multimodal}, which provides high-quality figure–caption pairs from peer-reviewed papers with heuristic filtering; and
RedPajama-ArXiv-Refined~\citep{datajuicer2023redpajama_arxiv_refined}, which supplies cleaned LaTeX source for extracting figure-associated context.
Intersecting these datasets by arXiv ID\footnote{The identifier assigned to each arXiv paper (e.g., 2301.12345).} yields 572K papers.
We randomly shuffle this pool to avoid temporal or domain-based ordering bias, then process papers in batches until reaching our target dataset size, ultimately processing 44{,}345 papers.

\noindent\textbf{Context Extraction.}
For each figure-caption pair \((F_i, C_i)\), we extract its associated context \(\mathcal{P}_i\) by identifying paragraphs that cite \(F_i\) via LaTeX reference commands (e.g., \verb|\ref|, \verb|\cref|, or \verb|\autoref|).
Details are provided in Appendix~\ref{app:data-prep}.

\noindent\textbf{Model Selection.}
We use \texttt{DeepSeek-v3}~\citep{liu2024deepseek} for \(M_{\text{text}}\) due to its strong instruction-following capability and wide adoption.
For distractor generation, we use \texttt{GPT-4o}~\citep{openai2024gpt4} as \(M_{\text{vision}}^{\text{gen}}\) for its balanced cost and overall capability.
For verification, we use \texttt{o4-mini}~\citep{openai_o4_mini_2025} as \(M_{\text{vision}}^{\text{verify}}\), which demonstrates strong scientific figure understanding on the CharXiv benchmark~\citep{wang2024charxiv}.
For the visual-dependence check, to save cost, QA pairs are first filtered by \texttt{DeepSeek-v3}, and only those that pass are further verified by \(M_{\text{vision}}^{\text{verify}}\). 
For vision-based filtering, we prompt \(M_{\text{vision}}^{\text{verify}}\) to answer the question, querying it three times with stochastic decoding (temperature=1.0) and applying 2-of-3 majority voting to retain only QA pairs where the model reliably selects \(A^*\).
Prompt templates for all stages are provided in Appendix~\ref{app:code-release}.


\noindent\textbf{Resulting Dataset.}
Applying our Cross-Modal Verification framework yields \ours, a dataset of 20{,}272 quality-controlled QA pairs spanning 20 arXiv categories. 
Each QA instance consists of a question, multiple-choice options, the correct answer, the associated figure, and its caption.
We release the full 20{,}272 QA pairs as a training resource; downstream effectiveness is evaluated exclusively on external benchmarks (\S\ref{subsec:training}).
For both training and evaluation, models receive the figure, its caption, the question, and the options as input, and are expected to predict the correct answer. This input design mirrors how readers naturally examine a scientific figure alongside its caption (our visual-dependence check helps ensure that the caption alone is insufficient to answer each question).
Table~\ref{tab:filtering-funnel} shows the filtering statistics at each step.

\begin{table}[t]
\centering
\footnotesize
\setlength{\tabcolsep}{4pt}
\caption{Number of QA pairs retained after each step.}\label{tab:filtering-funnel}
\begin{tabular}{l@{\hskip 6pt}r@{\hskip 6pt}r}
\toprule
Step & Count & Retention \\
\midrule
Papers & 44{,}345 & --- \\
Atomic claims extracted & 680{,}877 & 100\% \\
QA pairs generated & 261{,}116 & 38.4\% \\
After text-based filtering & 55{,}372 & 8.1\% \\
After vision-based filtering & 20{,}272 & 3.0\% \\
\bottomrule
\end{tabular}
\vspace{-0.1in}
\end{table}

The retention rate is low (3.0\%) because generating correct, visually grounded QA pairs from figure-citing paragraphs alone is inherently challenging, as confirmed by our small-scale case study of rejected QA pairs (Appendix~\ref{app:filtering-analysis}).
For example, claims in figure-citing paragraphs are often vague, making them unsuitable for generating QA pairs.
Our ablation study (\S\ref{subsec:ablation}) shows that this conservative design is effective, with each additional filtering step yielding clear downstream improvements. Human verification further confirms that the majority of rejected QA pairs are indeed erroneous (Appendix~\ref{app:filtering-analysis}).
Further dataset analysis is provided in Section~\ref{sec:exp}.

\section{Experiments}\label{sec:exp}

We first analyze the diversity of our dataset (\S\ref{subsec:diversity}), then evaluate leading LVLMs on it to verify that it is non-trivial for current models and to identify gaps between open-source and proprietary models that our dataset might help address (\S\ref{subsec:difficulty}), and compare our dataset's quality against existing SVQA datasets (\S\ref{subsec:quality}).
We show that fine-tuning on our dataset yields consistent gains across multiple external benchmarks, with the average gain scaling up with data size (\S\ref{subsec:training}), and verify the effectiveness of our filtering steps via ablation (\S\ref{subsec:ablation}).

\subsection{Diversity Assessment}\label{subsec:diversity}

\begin{figure*}[t]
\centering
\includegraphics[width=\textwidth]{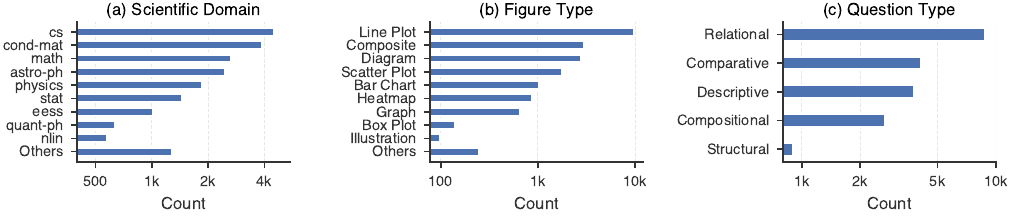}
\caption{
    Dataset composition of \ours across three dimensions: 
    (a)~scientific domain distribution showing the top 9 of 20 arXiv categories, 
    (b)~figure type coverage showing the top 9 of 12 categories, and 
    (c)~distribution across five question types.
    Complete breakdowns are in Appendix~\ref{app:taxonomy}.
}
\Description{Three bar charts showing dataset composition: (a) distribution across scientific domains with CS and Cond-Mat being the most frequent, (b) distribution across figure types with Line Plot and Composite being most common, and (c) distribution across question types with Relational and Comparative questions being most frequent.}
\label{fig:dataset-stats}
\end{figure*}

Our dataset comprises 20{,}272 diverse QA pairs spanning three key dimensions: scientific domains, figure types, and question types.
Scientific domain labels are assigned according to the arXiv category taxonomy~\cite{arxiv_taxonomy} of the source papers. Figure type and question type are automatically annotated using \texttt{GPT-4o}~\citep{openai2024gpt4}.
To verify the GPT-4o labels, we randomly sample 50 examples per figure-type and question-type category for human review; human agreement with GPT-4o labels is at least 96\% across all figure-type categories and at least 86\% across all question-type categories.
As shown in Fig.~\ref{fig:dataset-stats}, our dataset covers 20 scientific domains and 12 figure types. Since our data is sourced from arXiv, CS and Physics are the most frequent domains~\citep{li2024multimodal}; Line Plot is the most prevalent figure type, as plots and charts dominate scientific publications~\citep{pramanick2024spiqa}.
Notably, our dataset is reasoning-heavy: Relational, Comparative, and Compositional questions together comprise 77\% of the dataset. This is because our framework generates QA pairs from figure-citing paragraphs (\S\ref{subsec:generation}), which tend to convey key findings derived from the figure rather than low-level perceptual details.
Detailed taxonomy definitions and complete distribution tables are provided in Appendix~\ref{app:taxonomy}.

\subsection{Difficulty Assessment}
\label{subsec:difficulty}

We evaluate leading proprietary and open-source LVLMs on our dataset in a zero-shot setting.

\noindent\textbf{Setup.}
We evaluate six models on 2{,}000 randomly sampled examples: four proprietary models (\texttt{OpenAI o3}~\citep{openai_o3_2025}, \texttt{GPT-5.2}~\citep{openai_gpt52_2025}, \texttt{o4-mini}~\citep{openai_o4_mini_2025}, \texttt{Gemini-3-Flash}~\citep{gemini3flash}) and two open-source models (\texttt{Qwen3-VL-235B-A22B-Thinking}~\citep{qwen3vl}, \texttt{InternVL3.5-38B}~\citep{chen2024internvl}).
We use greedy decoding for all models except \texttt{Gemini-3-Flash} and \texttt{Qwen3-VL-235B-A22B-Thinking}, whose official documentation warns that greedy decoding may cause unexpected behavior such as endless output loops; for these two models, we adopt their recommended settings.
Detailed configurations are in Appendix~\ref{app:model-details}.

\begin{figure*}[t]
\centering
\includegraphics[width=\textwidth]{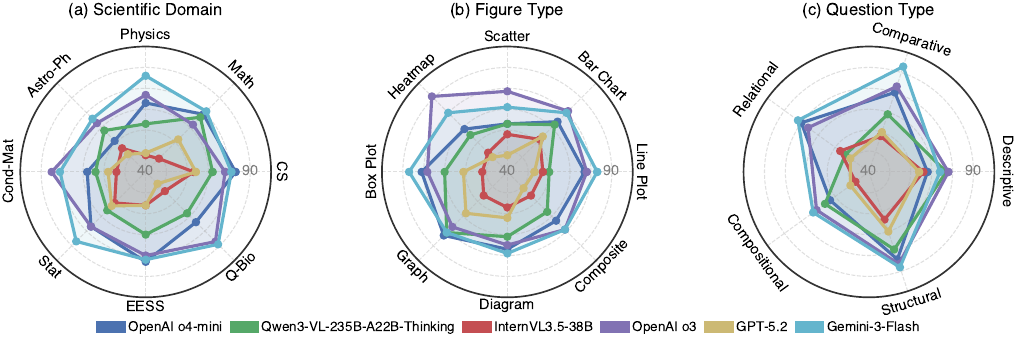}
\caption{
    Zero-shot model accuracy on \ours broken down by (a)~scientific domain, (b)~figure type, and (c)~question type.
    The largest gaps between proprietary and open-source models appear in domains like Physics, figure types like Box Plot and Heatmap, and question types like Comparative.
}
\Description{Three radar charts showing model performance: (a) accuracy across scientific domains with Astro-Ph and Physics being most challenging, (b) accuracy across figure types with Scatter Plot and Composite being hardest, and (c) accuracy across question types with Compositional and Relational questions being most difficult. Six models are compared: OpenAI o3, GPT-5.2, o4-mini, Gemini-3-Flash, Qwen3-VL-235B-A22B-Thinking, and InternVL3.5-38B.}
\label{fig:radar}
\vspace{-0.15in}
\end{figure*}

\noindent\textbf{Results.}
The best open-source model, \texttt{Qwen3-VL}, reaches 64.6\% overall accuracy.
The best proprietary model, \texttt{Gemini-3-Flash}, achieves 80.5\%, leaving considerable room for open-source models to catch up (detailed results in Appendix~\ref{app:detailed-results}).
As shown in Fig.~\ref{fig:radar}, category-level breakdowns reveal that the largest gaps appear in domains such as Physics and Q-Bio, figure types such as Box Plot, Line Plot, and Heatmap, and question types such as Comparative and Relational.
These challenging categories generally demand precise quantitative reading or multi-step reasoning, pinpointing specific weaknesses that our framework can help mitigate by synthesizing training data with greater coverage of the corresponding domains, figure types, and question types.
Beyond this gap analysis, we note that \texttt{Gemini-3-Flash} outperforms our verifier \texttt{o4-mini} on this sample, suggesting that the retained QA pairs are not trivially biased toward the verifier.

\subsection{Quality Comparison}
\label{subsec:quality}

We perform human evaluation on 600 QA pairs (200 each from our dataset, SPIQA~\citep{pramanick2024spiqa}, and ArXivQA~\citep{li2024multimodal}), sampled from the training split.
Ten CS graduate students with experience reading scientific papers serve as annotators. Before the formal annotation, all annotators studied detailed guidelines with anchor examples at each score level and were calibrated on 20 seed questions, proceeding only after reaching inter-annotator agreement (Krippendorff's $\alpha \geq 0.5$) on every dimension. Each QA pair is independently rated by two annotators.
We assess four quality dimensions corresponding to the error types in \S\ref{sec:motivation} on a 5-point Likert scale: Factual Correctness (E1), Intent Alignment (E2), Visual Dependency (E3), and Self-Containment (E4).
As shown in Table~\ref{tab:human-eval}, our dataset rates favorably on Factual Correctness, Intent Alignment, and Visual Dependency, while maintaining comparable Self-Containment.
Detailed evaluation protocol is in Appendix~\ref{app:human-eval}.

\begin{table}[t]
\centering
\caption{Human evaluation results on 200 QA pairs per dataset, sampled from training splits. Ratings are on a 5-point Likert scale; higher is better.}\label{tab:human-eval}
\footnotesize
\setlength{\tabcolsep}{4pt}
\begin{tabular}{lcccc}
\toprule
\multirow{2}{*}{Dimension} & \multicolumn{3}{c}{Rating (1--5)} & \multirow{2}{*}{IAA ($\alpha$)} \\
\cmidrule(lr){2-4}
 & \ours & SPIQA & ArXivQA & \\
\midrule
Factual Correctness  & \textbf{4.49} & 4.19 & 3.96 & 0.71 \\
Intent Alignment     & \textbf{4.35} & 4.34 & 4.06 & 0.62 \\
Visual Dependency    & \textbf{4.40} & 4.35 & 4.17 & 0.71 \\
Self-Containment     & 4.34 & \textbf{4.52} & 4.18 & 0.60 \\
\bottomrule
\end{tabular}
\vspace{-0.1in}
\end{table}

\noindent\textbf{Analysis.}
Our higher ratings on E1--E3 can be attributed to specific design choices: vision-based filtering for Factual Correctness (E1), generating questions and answers from figure-citing paragraphs for Intent Alignment (E2), and the visual-dependence check for Visual Dependency (E3).
For Self-Containment (E4), SPIQA scores higher.
Because our QA pairs are derived from figure-citing paragraphs, some QA pairs may rely on contextual details present in the paragraph but absent from the figure and caption.
Our vision-based filtering helps discard such outside-knowledge QA pairs (E4, \S\ref{sec:motivation}), but a few such pairs persist.
All four dimensions achieve moderate inter-annotator agreement ($\alpha$ = 0.60--0.71)~\citep{krippendorff2018content}.

\subsection{Effectiveness as Training Data}
\label{subsec:training}

\noindent\textbf{Setup.}
We compare the training effectiveness of our dataset against datasets representing different construction methods:
template-based SciFiBench~\citep{roberts2024scifibench} (Figure2Caption split, 1{,}000 examples from General and CS categories),
and LVLM-driven ArXivQA~\citep{li2024multimodal} and SPIQA~\citep{pramanick2024spiqa} (figure-type questions only), with 20{,}272 randomly sampled training examples each.
For our dataset, we train two variants: one using all 20{,}272 samples, and one using a 1{,}000-example subset with the same scientific domain distribution as SciFiBench for a controlled comparison at the same scale.
Using LoRA~\citep{hu2022lora} (rank=16, $\alpha$=32), we fine-tune \texttt{Qwen2.5-VL-7B-Instruct}~\citep{qwen2vl} for one epoch with a 9:1 train/validation split, selecting the best checkpoint based on validation performance.
All fine-tuned variants are evaluated on three external scientific figure benchmarks:
CharXiv~\citep{wang2024charxiv} (4{,}000 descriptive and 1{,}000 reasoning examples),
MMStar~\citep{chen2024we} (science \& technology subset, 250 examples), and MathVista~\citep{lu2023mathvista} (scientific reasoning subset, 122 examples).
Additional training details are in Appendix~\ref{app:training-details}.

\noindent\textbf{Results.}
We report performance gains relative to the original \texttt{Qwen2.5-VL-7B-Instruct} (baseline).
As shown in Table~\ref{tab:training-results}, the model trained on our dataset (20{,}272 examples) achieves the highest average gain (+2.05\%) compared to models trained on other datasets.
Notably, our dataset is the only one that improves CharXiv Reasoning (+2.10\%), whereas other datasets show degradation.
We attribute this improvement to our higher data quality (\S\ref{subsec:quality}) and reasoning-heavy question distribution (\S\ref{subsec:diversity}), as also reflected in the gains on MMStar Science \& Technology and MathVista Scientific Reasoning.
On CharXiv Descriptive, our dataset improves by +3.28\%, though ArXivQA leads with +5.42\%.

\begin{table}[t]
\centering
\footnotesize
\setlength{\tabcolsep}{4pt}
\caption{Absolute baseline accuracy (\%) and fine-tuning gains ($\Delta$\%). CX-D/R: CharXiv Descriptive/Reasoning; MM-ST: MMStar Science \& Technology; MV-SR: MathVista Scientific Reasoning. Bold indicates best gain per column.}
\label{tab:training-results}
\begin{tabular}{l@{\hskip 6pt}c@{\hskip 6pt}c@{\hskip 6pt}c@{\hskip 6pt}c@{\hskip 6pt}c}
\toprule
Training Data & CX-D & CX-R & MM-ST & MV-SR & Avg. \\
\midrule
Baseline (no fine-tuning) & 66.0 & 42.4 & 48.0 & 67.2 & 55.9 \\
\midrule
SciFiBench (1K) & +4.28 & $-$1.10 & +0.80 & 0.00 & +1.00 \\
\ours (1K)     & +5.50 & $-$1.00 & +0.80 & $-$0.82 & +1.12 \\
\midrule
ArXivQA (20K)   & \textbf{+5.42} & $-$1.40 & 0.00 & $-$0.81 & +0.80 \\
SPIQA (20K)     & +2.45 & $-$2.70 & +0.40 & $-$1.63 & $-$0.37 \\
\ours (20K)      & +3.28 & \textbf{+2.10} & \textbf{+2.00} & \textbf{+0.82} & \textbf{+2.05} \\
\bottomrule
\end{tabular}
\vspace{-0.05in}
\end{table}

\begin{figure}[!h]
\centering
\includegraphics[width=0.65\linewidth]{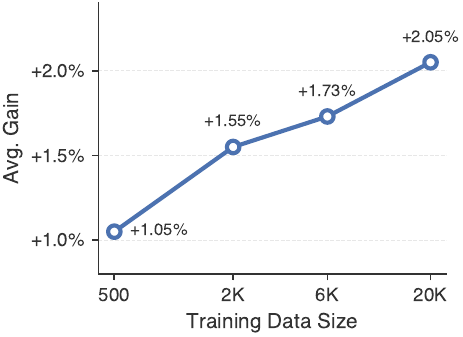}
\caption{Average gain with increasing amounts of our training data when fine-tuning \texttt{Qwen2.5-VL-7B-Instruct}.}
\Description{A line chart showing performance gains on the y-axis against training data size on the x-axis. The curve shows consistent improvement as training data scales from 500 to 20272 examples.}
\label{fig:scaling-curve}
\end{figure}

\noindent\textbf{Training Data Scaling.}
To assess the scalability of our dataset, we use the same training setup described above but vary the amount of training data from 500 to 20{,}272 examples.
As shown in Fig.~\ref{fig:scaling-curve}, the average gain scales monotonically with training data size, nearly doubling from +1.05\% to +2.05\% as the data grows from 500 to 20{,}272 examples.
This trend suggests potential for further improvement with more of our training data.
Scaling up our dataset is also practically feasible: on average, our framework requires approximately 59 DeepSeek-v3 calls, 13 GPT-4o calls, and 9 o4-mini calls to produce one verified QA pair; using publicly available APIs, we achieved a throughput of up to 1{,}200 verified QA pairs per day.

\subsection{Ablation Study}
\label{subsec:ablation}

\noindent\textbf{Setup.}
To assess the incremental contribution of each filtering step in our framework, we compare three variants, each trained on 20{,}272 QA pairs sampled at a different step of our framework (see Table~\ref{tab:filtering-funnel}): 
(1)~\textbf{Generation only}: QA pairs sampled directly after the Generation stage (\S\ref{subsec:generation}), with no filtering applied;
(2)~\textbf{With text-based filtering}: QA pairs sampled after applying text-based filtering (\S\ref{subsec:verification}), before vision-based filtering;
(3)~\textbf{With full verification}: all 20{,}272 QA pairs retained after both text-based and vision-based filtering (we directly reuse the result from \S\ref{subsec:training}).
Training and evaluation follow the same configuration as \S\ref{subsec:training}.

\begin{table}[t]
\centering
\footnotesize
\setlength{\tabcolsep}{6pt}
\caption{
    Average gains when training on 20K examples sampled at each step of our framework.
}\label{tab:ablation-results}
\vspace{-2mm}
\begin{tabular}{lc}
\toprule
Training Data (20K examples) & Avg. Gain (\%) \\
\midrule
Generation only (no filtering)     & $-$0.37 \\
With text-based filtering & +0.24 \\
With full verification     & \textbf{+2.05} \\
\bottomrule
\end{tabular}
\vspace{-0.1in}
\end{table}

\noindent\textbf{Results and Analysis.}
Training on unfiltered data hurts performance ($-$0.37\%), while text-based filtering provides a modest average gain of +0.24\%.
Although stopping at text-based filtering would retain 2.7$\times$ more QA pairs than full verification (see Table~\ref{tab:filtering-funnel}), this quantity advantage comes at a quality cost: without vision-based filtering, \emph{incorrectly visually grounded} (E1) and \emph{outside-knowledge} (E4) errors remain unfiltered.
Adding vision-based filtering boosts the average gain from +0.24\% to \textbf{+2.05\%} (8.5$\times$), confirming that cross-modal verification is essential even though it retains only about 36\% of the QA pairs that pass text-based filtering.
\vspace{-0.1in}

\section{Related Work}\label{sec:related}

\noindent\textbf{Scientific Visual Question Answering Datasets.}
Existing SVQA datasets range from template-based approaches~\cite{kahou2017figureqa,roberts2024scifibench,methani2020plotqa}, to expert annotation~\cite{wang2024charxiv,masry2022chartqa}, to LVLM-based generation~\cite{li2024multimodal,pramanick2024spiqa,shabtay2024livexiv,li2023scigraphqa}.
Among LVLM-generated datasets, the figure is always present during question generation, making outputs susceptible to visual hallucination. SPIQA~\cite{pramanick2024spiqa} mitigates information asymmetry by incorporating figure-citing paragraphs as context, but does not verify generated answers; LiveXiv~\cite{shabtay2024livexiv} introduces verification via inter-model agreement, yet both generation and verification operate on the same visual input, so correlated hallucinations may pass unfiltered.
In contrast, \ours generates questions and answers from figure-citing paragraphs without the figure, retaining the contextual benefit of these paragraphs, and then verifies against the figure itself. Because hallucinations introduced during text-based generation and those introduced during vision-based verification are less likely to be correlated, this cross-modal design helps catch erroneous QA pairs that same-figure verification might miss.

\noindent\textbf{Data Synthesis and Verification Techniques.}
Our methodology builds on recent advances in data synthesis and hallucination mitigation. 
Similar to self-instruction methods~\cite{wang2023self}, we leverage LLMs to generate data at scale.
Recent surveys on synthetic data~\cite{liu2024best,long2024llms} emphasize that data quality significantly outweighs quantity, and that systematic verification is essential for downstream model performance.
LiveXiv~\cite{shabtay2024livexiv} applies automated filtering by combining a ``blind test'', conceptually equivalent to our visual-dependence check (\S\ref{subsec:verification}), with agreement between disjoint models on the same figure.
However, as discussed in \S\ref{sec:intro}, agreement between LVLMs on the same figure may not reliably catch correlated hallucinations.
Our verification instead leverages separate modalities: text-generated answers are validated against the figure, exploiting the consistency between what authors write and what figures show. We apply cascaded text-based and vision-based checks (\S\ref{subsec:verification}), with the final vision-based step strengthened by self-consistency~\cite{wang2022self} via majority voting.
Unlike general fact-checking methods that rely on external knowledge bases~\cite{lewis2020retrieval,nakano2021webgpt,gao2023rarr}, our verification is grounded in the paper's own context and the figure, requiring no external retrieval.

\noindent\textbf{Large Vision-Language Models for Science.}
The success of recent LVLMs~\cite{alayrac2022flamingo,liu2023visual,OpenAI_GPT4V_SystemCard_2023,chen2024internvl} has spurred interest in their application to scientific domains. 
Specialized models for chart understanding include MatCha~\cite{liu2023matcha}, which enhances visual pretraining with math reasoning, ChartLlama~\cite{han2023chartllama} for diverse chart tasks, and TinyChart~\cite{zhang2024tinychart}, which achieves strong performance with only 3B parameters through program-of-thought learning.
Benchmarks such as CharXiv~\cite{wang2024charxiv}, SciFiBench~\cite{roberts2024scifibench}, ChartX~\cite{he2024chartx}, and MathVista~\cite{lu2023mathvista} evaluate LVLMs' capabilities on scientific figures and mathematical reasoning.
Beyond scientific domains, general LVLM benchmarks such as MMBench~\cite{liu2024mmbench}, MME~\cite{fu2025mme}, and MMStar~\cite{chen2024we} evaluate broad multimodal capabilities.
These evaluations consistently reveal a significant capability gap: while proprietary models demonstrate strong performance, open-source counterparts lag behind in scientific visual reasoning.
Training datasets for SVQA remain limited in scale or quality. We propose a framework for constructing large-scale, verified data as one approach to help bridge this gap.
\vspace{-0.1in}
\section{Conclusion}
We curate \ours, a dataset of 20{,}272 high-quality multiple-choice SVQA pairs spanning 20 scientific domains, created via our Cross-Modal Verification framework that produces reliable training data from unreliable generators by leveraging cross-modal consistency rather than expensive human annotation. Experiments show that fine-tuning on \ours yields consistent gains across SVQA benchmarks, with improvements scaling with data size. We hope this work can extend beyond SVQA to other domains where textual descriptions align with visual content, such as medical imaging and technical documentation, and inspire further research on cross-modal data verification.

\section*{Use of Generative AI}

In accordance with the ACM Policy on Authorship, we disclose the use of generative AI in this work.

\noindent\textbf{As a Research Object.}
Generative AI models are central to our proposed Cross-Modal Verification framework for dataset curation.
Specifically, we use DeepSeek-v3~\citep{liu2024deepseek} for atomic claim extraction, question--answer generation, and text-based filtering;
GPT-4o~\citep{openai2024gpt4} for visually grounded distractor generation;
and o4-mini~\citep{openai_o4_mini_2025} for vision-based verification.
These models serve as core components of our methodology and their usage is fully described in \S\ref{sec:method}.

\noindent\textbf{Writing Assistance.}
We used ChatGPT to assist with grammar checking and polishing of the manuscript text.
All generated content was reviewed and revised by the authors, who take full responsibility for the final manuscript.


\bibliographystyle{ACM-Reference-Format}
\bibliography{main}

\appendix

\noindent\textbf{Appendix Overview.}\vspace{2pt}
\begingroup\small
\appsecentry{Implementation Details}{app:implementation}
\appsubentry{Data Preparation}{app:data-prep}
\appsubentry{Generation}{app:generation}
\appsubentry{Verification}{app:verification}
\vspace{2pt}
\vspace{2pt}
\appsecentry{Dataset Taxonomies}{app:taxonomy}
\appsubentry{Figure Type Taxonomy}{app:figure-taxonomy}
\appsubentry{Question Type Taxonomy}{app:question-taxonomy}
\appsubentry{Scientific Domain Taxonomy}{app:domain-taxonomy}
\vspace{2pt}
\appsecentry{Filtering Analysis}{app:filtering-analysis}
\appsubentry{QA Pair Generation Analysis}{app:qa-gen-analysis}
\appsubentry{Text-Based Filtering Analysis}{app:text-filter-analysis}
\appsubentry{Vision-Based Filtering Analysis}{app:vision-filter-analysis}
\vspace{2pt}
\appsecentry{Human Evaluation Protocol}{app:human-eval}
\appsubentry{Annotation Interface}{app:annotation-interface}
\appsubentry{Annotation Guidelines}{app:annotation-guidelines}
\vspace{2pt}
\appsecentry{Additional Qualitative Examples}{app:examples}
\appsubentry{QA Pairs in \ours}{app:qa-examples}
\appsubentry{Error Cases Captured by Framework}{app:error-examples}
\vspace{2pt}
\appsecentry{Model Details}{app:model-details}
\vspace{2pt}
\appsecentry{Per-Category Performance Results}{app:detailed-results}
\appsubentry{Performance by Scientific Domain}{app:perf-domain}
\appsubentry{Performance by Figure Type}{app:perf-figtype}
\appsubentry{Performance by Question Type}{app:perf-questype}
\vspace{2pt}
\appsecentry{Training Details}{app:training-details}
\vspace{2pt}
\appsecentry{Framework Instantiation with Data-Juicer}{app:data-pipeline}
\appsubentry{Custom Operator: LaTeX Figure Context Extractor}{app:custom-operator}
\appsubentry{Custom Operator: Claim Extractor}{app:custom-operator-claim}
\appsubentry{Operator Orchestration}{app:pipeline-orchestration}
\appsubentry{Code Release}{app:code-release}
\endgroup

\section{Implementation Details}\label{app:implementation}

This section provides technical details about our Cross-Modal Verification framework instantiation to curate \ours, covering data preparation, generation, and verification stages.
A reusable Data-Juicer implementation is detailed in Appendix~\ref{app:data-pipeline}.

\subsection{Data Preparation}
\label{app:data-prep}

Our framework requires figure-associated context \(\mathcal{P}\) (paragraphs that cite the figure) to generate QA pairs via atomic claim extraction (Eq.~\ref{eq:claim-extraction}), but existing datasets like ArXivCap~\citep{li2024multimodal} provide only figure-caption pairs \((F, C)\).
We therefore augment ArXivCap to construct triplets \((F, C, \mathcal{P})\), where \(\mathcal{P}\) consists of paragraphs that explicitly cite figure \(F\).
We detail this process below.

\noindent\textbf{Data Sources.}
We use two data sources: ArXivCap\footnote{\url{https://huggingface.co/datasets/MMInstruction/ArxivCap}}~\citep{li2024multimodal} for figure-caption pairs \((F, C)\) and RedPajama-ArXiv-Refined\footnote{\url{https://huggingface.co/datasets/datajuicer/redpajama-arxiv-refined-by-data-juicer}}~\citep{datajuicer2023redpajama_arxiv_refined} for LaTeX source files.
Matching by arXiv IDs yields 572K overlapping papers.
For each matched paper, we bind its LaTeX source to the figure-caption pairs from ArXivCap.
We randomly shuffle this pool and process 44{,}345 papers in batches to reach our target dataset size.

\noindent\textbf{Context Extraction.}
For each figure-caption pair $(F,C)$ in a matched paper, we extract its associated context $\mathcal{P}$ from the LaTeX source as follows.
We first associate figure $F$ with its corresponding figure environment by matching captions.
Since ArXivCap normalizes captions using \texttt{pylatexenc},\footnote{\url{https://github.com/phfaist/pylatexenc}} we apply the same normalization to enable efficient matching via Levenshtein similarity~\citep{levenshtein1966binary} ($\text{sim} \geq 0.9$).
We then use regular expressions to extract figure labels from \verb|\label| commands within matched figure environments (including subfigure environments).
Figures with empty captions, ambiguous matches (e.g., duplicate captions), or those lacking a valid label are discarded.
We then identify all paragraphs that cite figure $F$ by searching for the extracted label in \verb|\ref|, \verb|\cref|, and \verb|\autoref| commands.\looseness=-1
For figures cited in multiple paragraphs, we retain and concatenate all citing paragraphs as context~$\mathcal{P}$.

\subsection{Generation}
\label{app:generation}

Given triplets \((F, C, \mathcal{P})\) from data preparation, this stage generates candidate QA pairs through three steps: claim extraction (Eq.~\ref{eq:claim-extraction}), question \& answer generation (Eq.~\ref{eq:qa-generation}), and distractor generation (Eq.~\ref{eq:distractor-generation}).
We use \texttt{deepseek-v3-0324}~\citep{liu2024deepseek} as \(M_{\text{text}}\) with temperature \(T=1.0\).
All prompts are constructed using Jinja2~\citep{jinja2} templates; complete templates are provided in Appendix~\ref{app:code-release}.

\noindent\textbf{Claim Extraction (Eq.~\ref{eq:claim-extraction}).}
We instruct $M_{\text{text}}$ to extract atomic claims $\mathcal{S} = \{s_j\}$ from context $\mathcal{P}$, outputting them within XML tags, with each claim $s_j$ following the pattern ``The figure shows...''.
If no valid claim can be extracted, the model outputs ``None'', and such responses are discarded.

\noindent\textbf{Question \& Answer Generation (Eq.~\ref{eq:qa-generation}).}
For each atomic claim \(s_j \in \mathcal{S}\), we prompt \(M_{\text{text}}\) to generate a question \(Q_j\) and its correct answer \(A^*_j\) in XML format.
The model may output ``None'' if \(s_j\) lacks sufficient detail; such cases are discarded.

\noindent\textbf{Distractor Generation (Eq.~\ref{eq:distractor-generation}).}
To complete the multiple-choice format, we prompt \(M_{\text{vision}}^{\text{gen}}\) (\texttt{gpt-4o-2024-08-06}~\citep{openai2024gpt4}) with figure \(F\), caption \(C\), question \(Q_j\), and correct answer \(A^*_j\) to generate three plausible distractor options, yielding the full option set \(O_j\) and candidate QA pair \(q_j = (Q_j, O_j, A^*_j)\).

\subsection{Verification}
\label{app:verification}

Given candidate QA pairs from generation, this stage applies cascaded filtering (Eq.~\ref{eq:passcriterion}) to retain only high-quality instances.
Text-based filtering first ensures source-con\-sis\-ten\-cy and visual-de\-pen\-dence without accessing the figure; vision-based filtering then validates each remaining QA candidate against the figure.
We use \texttt{o4-mini-2025-04-16}~\citep{openai_o4_mini_2025} as $M_{\text{vision}}^{\text{verify}}$ with $T{=}1.0$.

\noindent\textbf{Text-Based Filtering.}
The first step applies two figure-free filters without accessing figure \(F\):

\noindent\textbf{Source-Consistency Check (\(V_{\text{src}}\)).}
We prompt \(M_{\text{text}}\) to answer question \(Q_j\) from options \(O_j\) based solely on source context \(\mathcal{P}\).
The model outputs its selected answer in an \verb|<option>| tag, or ``None'' if the information is insufficient to determine a unique answer.
The filter \(V_{\text{src}}(Q_j, O_j, A^*_j, \mathcal{P})\) returns \textbf{True} only if the model's selection exactly matches \(A^*_j\), and \textbf{False} otherwise (including ``None'', multiple answers, or cases where a valid answer cannot be extracted).

\noindent\textbf{Visual-Dependence Check (\(V_{\text{vis\_dep}}\)).}
To reduce computational cost, we implement this check in two cascaded steps.
First, QA pairs are evaluated by \texttt{deepseek-v3-0324} given only \(Q_j\), \(O_j\), and caption \(C\); only candidates where it fails to select \(A^*_j\) proceed.
Second, remaining candidates are validated by \(M_{\text{vision}}^{\text{verify}}\).
The filter \(V_{\text{vis\_dep}}(Q_j, O_j, A^*_j, C)\) returns \textbf{True} only if \(M_{\text{vision}}^{\text{verify}}\) also fails to select \(A^*_j\), indicating that \(Q_j\) requires visual information from figure \(F\) beyond caption \(C\).

\noindent\textbf{Vision-Based Filtering (\(V_{\text{vis\_con}}\)).}
We prompt \(M_{\text{vision}}^{\text{verify}}\) with figure \(F\), caption \(C\), question \(Q_j\), and options \(O_j\) to select an answer with reasoning.
We perform 3 independent queries and apply 2-of-3 majority voting; candidates with ties are discarded.
The filter \(V_{\text{vis\_con}}(Q_j, O_j, A^*_j, F, C)\) returns \textbf{True} only if the majority-voted option matches \(A^*_j\).
For retained QA pairs, we store the reasoning from the first query that agrees with the majority answer.


\section{Dataset Taxonomies}
\label{app:taxonomy}

This section provides detailed definitions and complete distribution tables for the three key dimensions of \ours : figure types, question types, and scientific domains. Scientific domain labels are assigned according to the arXiv category taxonomy of the source papers, while figure types and question types are automatically annotated using \texttt{GPT-4o} with carefully designed prompts (see Appendix~\ref{app:code-release}).

\subsection{Figure Type Taxonomy}
\label{app:figure-taxonomy}

We categorize figures in our dataset into 12 distinct types based on their visual structure and information encoding. This diversity helps models generalize across different visualization modalities and enables fine-grained performance analysis.

\noindent\textbf{Figure Type Definitions.}
Our figure taxonomy encompasses both data visualization charts and structural representations:

\begin{itemize}[leftmargin=*, itemsep=2pt, topsep=4pt]
    \item \textbf{Line Plot} (47.7\%): Graphs with continuous lines showing trends, relationships, or functions over continuous variables. Examples include accuracy curves over epochs, loss curves during training, and time series with confidence intervals.
    
    \item \textbf{Composite} (14.7\%): Figures containing multiple subplots/panels where different subplots belong to different visualization types (e.g., combining line plots with bar charts, or mixing scatter plots with heatmaps). Note: Figures with multiple subplots of the same type are classified into that specific type, not as Composite.
    
    \item \textbf{Diagram} (13.8\%): Schematic representations including neural network architectures, system pipelines, algorithm flowcharts, geometric diagrams, and physics/engineering schematics with technical annotations.
    
    \item \textbf{Scatter Plot} (8.8\%): Individual data points plotted without connecting lines, showing correlations, distributions, or clusters. Examples include t-SNE visualizations and feature space distributions.
    
    \item \textbf{Bar Chart} (5.1\%): Rectangular bars comparing discrete values across categories, commonly used for method performance comparison and ablation study results.
    
    \item \textbf{Heatmap} (4.3\%): Color-coded grids or matrices showing values through color intensity, including confusion matrices, attention maps, and correlation matrices.
    
    \item \textbf{Graph} (3.2\%): Graph/network visualizations showing nodes and edges as graph-theoretic structures, including knowledge graphs and dependency graphs.
    
    \item \textbf{Box Plot} (0.7\%): Statistical plots showing data distributions through quartiles, comparing distributions across categories.
    
    \item \textbf{Other} (0.5\%): Unusual chart types not covered by standard categories, such as radar charts, parallel coordinates, and Sankey diagrams.
    
    \item \textbf{Illustration} (0.5\%): Stylized drawings or artistic renderings, including computer-generated scenes and conceptual visualizations.
    
    \item \textbf{Photo} (0.5\%): Real photographs captured by cameras, typically used as dataset samples or experimental inputs.
    
    \item \textbf{Pie Chart} (0.2\%): Circular charts divided into slices showing proportions or percentages of categorical data.
\end{itemize}

The complete distribution is shown in Table~\ref{tab:figure-type-dist}. Line plots dominate the dataset (47.7\%), reflecting their prevalence in scientific publications for showing experimental results and trends. Composite figures (14.7\%) and diagrams (13.8\%) are also common, highlighting the complexity of scientific communication that often requires multiple visualization modalities or structural explanations.

\begin{table}[h]
\centering
\small
\begin{tabular}{lrr}
\toprule
\textbf{Figure Type} & \textbf{Count} & \textbf{Percentage} \\
\midrule
Line Plot & 9,662 & 47.7\% \\
Composite & 2,971 & 14.7\% \\
Diagram & 2,803 & 13.8\% \\
Scatter Plot & 1,787 & 8.8\% \\
Bar Chart & 1,031 & 5.1\% \\
Heatmap & 879 & 4.3\% \\
Graph & 655 & 3.2\% \\
Box Plot & 139 & 0.7\% \\
Other & 111 & 0.5\% \\
Illustration & 98 & 0.5\% \\
Photo & 95 & 0.5\% \\
Pie Chart & 41 & 0.2\% \\
\midrule
\textbf{Total} & \textbf{20,272} & \textbf{100.0\%} \\
\bottomrule
\end{tabular}
\caption{Distribution of figure types in our dataset. The taxonomy covers 12 distinct visualization modalities, from standard data plots (Line Plot, Bar Chart) to structural representations (Diagram, Graph) and composite multi-panel figures.}
\label{tab:figure-type-dist}
\end{table}

\subsection{Question Type Taxonomy}
\label{app:question-taxonomy}

We define 5 question types based on the cognitive operations required to answer them. This diversity helps models generalize across different reasoning operations and enables fine-grained performance analysis.

\noindent\textbf{Question Type Definitions.}
Our cognitive taxonomy categorizes questions by the primary reasoning operation required:

\begin{itemize}[leftmargin=*, itemsep=2pt, topsep=4pt]
    \item \textbf{Relational} (43.5\%): Identifying trends, patterns, correlations, states, or relationships between variables. Examples: ``Does the loss curve show convergence?''; ``What is the relationship between learning rate and accuracy?''; ``Which curve corresponds to the fluid state?''
    
    \item \textbf{Comparative} (20.2\%): Comparing two or more entities based on quantitative or qualitative measures (magnitude, quality, ranking). Examples: ``Which model performs best?''; ``Is Model A better than Model B?''; ``Which dataset shows the highest accuracy?''
    
    \item \textbf{Descriptive} (18.7\%): Reading or identifying a single value, label, or attribute directly from the figure. Examples: ``What is the accuracy of Model A?''; ``Which color represents the baseline?''; ``What is the value at epoch 10?''
    
    \item \textbf{Compositional} (13.2\%): Aggregation, computation, or multi-step reasoning involving multiple values. Examples: ``What is the average of values A, B, and C?''; ``What is the sum of all baseline scores?''; ``How many models are shown in the legend?''
    
    \item \textbf{Structural} (4.4\%): Understanding figure organization, architecture, flow, or structural relationships. Examples: ``What component follows the attention layer?''; ``How many stages are in the pipeline?''; ``What is the overall architecture?''
\end{itemize}

The complete distribution is shown in Table~\ref{tab:question-type-dist}. Relational questions dominate (43.5\%), indicating that scientific figure understanding heavily relies on pattern recognition and relationship analysis rather than simple value extraction. This distribution reflects the analytical nature of scientific inquiry, where understanding trends and correlations is more valuable than reading individual data points.

\begin{table}[h]
\centering
\small
\begin{tabular}{lrr}
\toprule
\textbf{Question Type} & \textbf{Count} & \textbf{Percentage} \\
\midrule
Relational & 8,811 & 43.5\% \\
Comparative & 4,087 & 20.2\% \\
Descriptive & 3,792 & 18.7\% \\
Compositional & 2,683 & 13.2\% \\
Structural & 899 & 4.4\% \\
\midrule
\textbf{Total} & \textbf{20,272} & \textbf{100.0\%} \\
\bottomrule
\end{tabular}
\caption{Distribution of question types in our dataset based on cognitive operations. The taxonomy emphasizes higher-order reasoning skills (Relational, Comparative, Compositional) over simple value extraction (Descriptive).}
\label{tab:question-type-dist}
\end{table}

\subsection{Scientific Domain Taxonomy}
\label{app:domain-taxonomy}

Our dataset spans 20 arXiv primary categories, covering diverse scientific fields from computer science and physics to biology and economics. This broad coverage helps models generalize across scientific domains with varying visual conventions and terminology, and enables fine-grained performance analysis.

\noindent\textbf{Domain Coverage.}
The distribution across scientific domains is shown in Table~\ref{tab:category-dist}. Computer Science (cs, 22.3\%) and Condensed Matter Physics (cond-mat, 19.1\%) are the most represented categories, followed by Mathematics (math, 13.0\%) and Astrophysics (astro-ph, 12.1\%). This distribution reflects both the prevalence of these fields on arXiv and their heavy reliance on visual figures for communicating results. The dataset also includes substantial representation from applied sciences (Statistics, Electrical Engineering, Quantitative Biology) and physical sciences (high-energy physics, nuclear physics, quantum physics), ensuring broad domain coverage.

\begin{table}[h]
\centering
\footnotesize
\setlength{\tabcolsep}{3pt}
\begin{tabular}{p{6cm}rr}
\toprule
\textbf{ArXiv Category} & \textbf{Count} & \textbf{\%} \\
\midrule
cs (Computer Science) & 4,519 & 22.3 \\
cond-mat (Condensed Matter) & 3,862 & 19.1 \\
math (Mathematics) & 2,634 & 13.0 \\
astro-ph (Astrophysics) & 2,455 & 12.1 \\
physics (Physics) & 1,849 & 9.1 \\
stat (Statistics) & 1,443 & 7.1 \\
eess (Electrical Engineering and Systems Science) & 1,012 & 5.0 \\
quant-ph (Quantum Physics) & 639 & 3.2 \\
nlin (Nonlinear Sciences) & 576 & 2.8 \\
q-bio (Quantitative Biology) & 526 & 2.6 \\
hep-ph (High Energy Physics -- Phenomenology) & 217 & 1.1 \\
nucl-th (Nuclear Theory) & 124 & 0.6 \\
gr-qc (General Relativity and Quantum Cosmology) & 104 & 0.5 \\
hep-th (High Energy Physics -- Theory) & 75 & 0.4 \\
q-fin (Quantitative Finance) & 68 & 0.3 \\
nucl-ex (Nuclear Experiment) & 49 & 0.2 \\
math-ph (Mathematical Physics) & 45 & 0.2 \\
hep-lat (High Energy Physics -- Lattice) & 26 & 0.1 \\
hep-ex (High Energy Physics -- Experiment) & 26 & 0.1 \\
econ (Economics) & 23 & 0.1 \\
\midrule
\textbf{Total} & \textbf{20,272} & \textbf{100.0} \\
\bottomrule
\end{tabular}
\caption{Distribution of scientific domains (arXiv primary categories) in our dataset. The broad coverage across 20 categories ensures domain diversity and generalization capability.}
\label{tab:category-dist}
\end{table}


\section{Filtering Analysis}\label{app:filtering-analysis}

In \S\ref{sec:exp}, we demonstrated that our Cross-Modal Verification framework produces higher-quality QA pairs compared to existing datasets through both human evaluation (\S\ref{subsec:quality}) and training effectiveness (\S\ref{subsec:training}).
To validate the effectiveness of each step, we conduct a small-scale human evaluation on rejected instances.
Results suggest the framework effectively identifies and removes the error types (E1--E4) identified in \S\ref{sec:motivation}.

\subsection{QA Pair Generation Analysis}\label{app:qa-gen-analysis}

During the QA pair generation step, \(M_{\text{text}}\) was instructed to decline producing QA pairs from atomic claims that are too vague or lack sufficient grounding (as described in \S\ref{subsec:generation}). To evaluate whether this rejection mechanism works as expected, we randomly sampled 30 rejected claims from the dataset curation process and manually evaluated them.

\noindent\textbf{Evaluation Results.} Among the 30 sampled rejected claims:
\begin{itemize}
    \item \textbf{Good (7 claims, 23.3\%)}: These atomic claims were actually suitable for QA generation but rejected due to \(M_{\text{text}}\) limitations. However, given the large pool of atomic claims available, this does not hinder the scalability of the framework.
    \item \textbf{Too Vague (23 claims, 76.7\%)}: These atomic claims lacked sufficient grounding to generate valid QA pairs. They either lacked specific visual elements (e.g., data points, labels, structural relationships), or used subjective qualitative descriptors without quantitative grounding (e.g., ``good results'', ``nearly indistinguishable''), making it difficult to construct objective, verifiable questions. See qualitative examples in Appendix~\ref{app:examples}.
\end{itemize}

\noindent\textbf{Discussion.} The analysis reveals that the generation refusal mechanism effectively filters unsuitable claims, with approximately 76.7\% of rejections being justified. While 23.3\% of rejections are due to \(M_{\text{text}}\) limitations, the large pool of atomic claims ensures this does not impact the framework's ability to scale.
The ``too vague'' cases can be attributed to authors' writing style in scientific papers, where figure descriptions are often briefly mentioned or glossed over in the main text without providing sufficient detail. When authors reference figures, they frequently use high-level or qualitative descriptions, assuming readers will examine the figure directly for specifics.

\subsection{Text-Based Filtering Analysis}\label{app:text-filter-analysis}

As described in \S\ref{subsec:verification}, text-based filtering applies two checks: source-consistency and visual-dependence. To verify these filters work as intended, we randomly sampled 50 QA pairs rejected after each step and manually evaluated them.

\noindent\textbf{Evaluation Results.} For the 50 sampled instances rejected by the source-consistency check:
\begin{itemize}
    \item \textbf{Valid rejection (36 instances, 72.0\%)}: \(M_{\text{text}}\) was unable to uniquely identify one correct answer from the provided options based on the source context, indicating genuine inconsistencies or ambiguities in the generated QA pairs.
    \item \textbf{Over-filtering (14 instances, 28.0\%)}: Valid QA pairs that were rejected due to \(M_{\text{text}}\) limitations or overly aggressive prompting, but should have been retained.
\end{itemize}

\noindent For the 50 sampled instances rejected by the visual-dependence check:
\begin{itemize}
    \item \textbf{Valid rejection (33 instances, 66.0\%)}: Questions that can be answered or reasonably inferred without viewing the figure, either from the caption alone or from general knowledge, correctly identifying non-visual questions (E3). Notably, some questions can be answered by reasoning through the answer options, which suggests that constructing more challenging distractor options could further improve question difficulty and visual dependence.
    \item \textbf{Over-filtering (17 instances, 34.0\%)}: Valid visual questions that were incorrectly rejected. When we instruct \(M_{\text{text}}\) to attempt answering the question given only the caption, \(M_{\text{text}}\) sometimes makes lucky guesses on questions that genuinely require visual information, accidentally matching the correct answer \(A^*\). This causes the filter to mistakenly classify these visual-dependent questions as answerable without the figure, leading to false rejections.
\end{itemize}

\noindent\textbf{Discussion.} The text-based filtering achieves 72.0\% and 66.0\% precision for source-consistency and visual-dependence checks respectively. The over-filtering rates (28.0\% and 34.0\%) reflect trade-offs in the filtering design: for source-consistency, aggressive prompting helps catch genuine inconsistencies but may be overly strict; for visual-dependence, \(M_{\text{text}}\) sometimes makes lucky guesses on genuinely visual questions. These results suggest that the text-based filtering works as intended.

\subsection{Vision-Based Filtering Analysis}\label{app:vision-filter-analysis}

As described in \S\ref{subsec:verification}, vision-based filtering validates that the answer is visually grounded in the figure and does not require outside knowledge, primarily targeting \emph{incorrectly visually grounded} (E1) and \emph{outside-knowledge} (E4) errors. To verify this filter works as intended, we randomly sampled 50 QA pairs rejected at the vision-based filtering and manually evaluated them.

\noindent\textbf{Evaluation Results.} The breakdown of the 50 sampled rejected instances is as follows:
\begin{itemize}
    \item \textbf{Incorrectly visually grounded (E1) (25 instances, 50.0\%):} The generated questions reference visual elements that are not present or not correctly represented in the figure. In most cases, $M_{\text{text}}$ hallucinates details when generating questions from atomic claims. In the remaining cases, the atomic claim itself contains information that does not correspond to the visual content, representing author errors. See Figure~\ref{fig:error_vis_e1} in Appendix~\ref{app:examples} for an example.
    
    \item \textbf{Outside-knowledge required (E4) (16 instances, 32.0\%):} The questions require information not visually available in the figure or caption. This error stems from \(M_{\text{text}}\) generating QA pairs conditioned on figure-associated context \(\mathcal{P}\), which may include information not present in the figure (e.g., precise numerical values, experimental settings, or quantitative details only mentioned in surrounding text). See Figure~\ref{fig:error_vis_e4_1} in Appendix~\ref{app:examples} for an example.
    
    \item \textbf{Over-filtering due to model limitations (9~instances, 18.0\%):} Valid QA pairs where the answer is visually grounded and answerable from the figure, but the model failed to provide the correct response due to its capability limitations.
\end{itemize}

\noindent\textbf{Discussion.} The vision-based filtering demonstrates strong precision (82.0\%) in identifying genuinely problematic QA pairs, successfully catching E1 and E4 errors that slipped through text-based filtering. The relatively low over-filtering rate (18.0\%) suggests that aggressive vision-based filtering is justified, as it removes 82\% problematic cases while only sacrificing 18\% valid instances due to \(M_{\text{vision}}^{\text{verify}}\) limitations. Combined with the ablation study results (\S\ref{subsec:ablation}) showing 8.5$\times$ performance gains from vision-based filtering, these findings suggest that ensuring visual grounding correctness and self-containment (not requiring outside knowledge) plays a critical role in dataset quality.


\section{Human Evaluation Protocol}\label{app:human-eval}

We describe the interface and annotation guidelines used in our human evaluation study (\S~\ref{subsec:quality}). We recruited 10 annotators who are CS graduate students with experience reading scientific papers. To ensure annotation quality, annotators were first asked to read detailed guidelines with concrete anchor examples at each score level (1, 3, 5). They then completed a calibration round on 20 seed questions; annotators proceeded to the formal annotation only after achieving inter-annotator agreement (Krippendorff's $\alpha \geq 0.5$) on every evaluation dimension.

\subsection{Annotation Interface}\label{app:annotation-interface}
\begin{figure}[h]
\centering
\includegraphics[width=\linewidth]{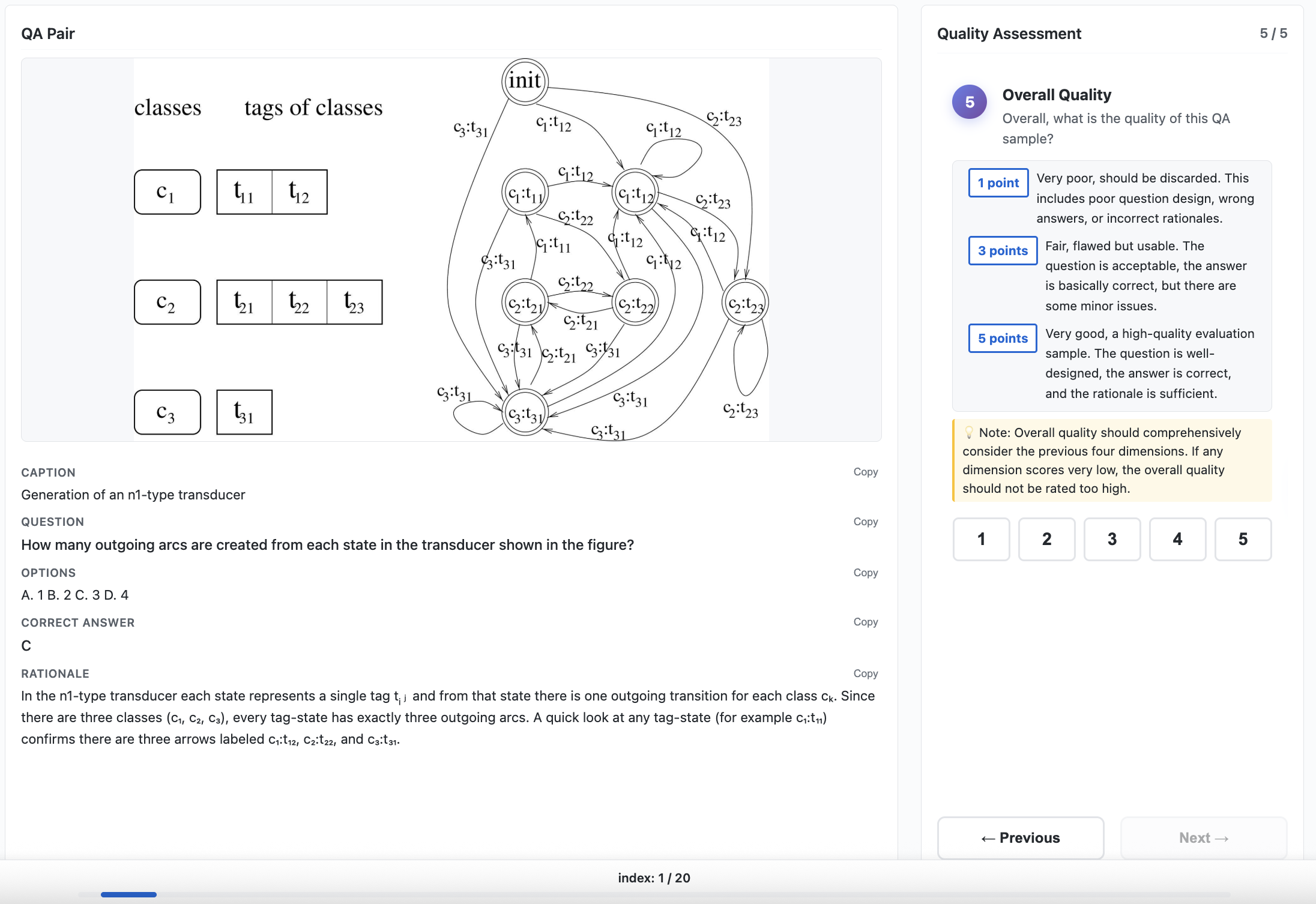}
\caption{Screenshot of the web-based annotation interface. The left panel displays the question information (figure, caption, question, options, correct answer, and rationale), while the right panel presents a carousel of rating cards for the five evaluation dimensions. Annotators rate each dimension on a 5-point Likert scale before submitting their assessment.}
\label{fig:annotation-interface}
\end{figure}

We developed a web-based annotation interface that presents annotators with a figure \(F\), caption \(C\), question \(Q\), answer options \(O\), and the designated correct answer \(A^*\) (following the notation from \S~\ref{subsec:formalization}). For QA pairs without captions (e.g., samples from baseline datasets like ArXivQA), the caption field is not displayed. Annotators were not provided with any background information about the datasets or our project to ensure unbiased evaluation. They evaluate each QA pair along multiple dimensions based solely on the presented information. Figure~\ref{fig:annotation-interface} shows a screenshot of the interface.

\subsection{Annotation Guidelines}
\label{app:annotation-guidelines}

We conducted human evaluation to assess the quality of our dataset across five key dimensions. This subsection presents the detailed annotation guidelines provided to human annotators.

\noindent\textbf{Overview.}
For each VQA sample, annotators were shown the figure \(F\), caption \(C\), question \(Q\), answer options \(O\), correct answer \(A^*\), and rationale. They were asked to rate the sample on five dimensions using a 5-point Likert scale (1 = lowest quality, 5 = highest quality). For each dimension, we provide anchor descriptions at scores 1, 3, and 5; annotators may assign intermediate scores (2 or 4) for cases falling between these anchors.

\medskip
\noindent\textbf{Dimension 1: Factual Correctness}

\noindent\textbf{Definition.} To what extent is the question, answer, and rationale explicitly supported by the visual information in the figure?

\noindent\textbf{Rating Criteria:}
\begin{itemize}[leftmargin=*, itemsep=2pt]
    \item \textbf{1 point}: Completely incorrect or contradicts the figure. This includes wrong answers, completely incorrect rationales, or hallucinated question content (asking about elements not present in the figure).
    \item \textbf{3 points}: Ambiguous or partially correct. The answer may be correct but the rationale is insufficient, or the answer might be correct but the evidence is unclear.
    \item \textbf{5 points}: Completely correct. The question is based on actual figure content, the answer is accurate, and the rationale has clear visual evidence support.
\end{itemize}

\noindent\textbf{Note:} Question hallucination (asking about content not present in the figure) counts as factually incorrect and should be rated 1 point.

\medskip
\noindent\textbf{Dimension 2: Intent Alignment}

\noindent\textbf{Definition.} To what extent does the question reflect the core scientific information that the figure is intended to convey?

\noindent\textbf{Rating Criteria:}
\begin{itemize}[leftmargin=*, itemsep=2pt]
    \item \textbf{1 point}: Completely irrelevant, such as treating a schematic diagram as a landscape photo. The question completely misses the scientific intent.
    \item \textbf{3 points}: Somewhat off-target but still within a reasonable range. The question asks about secondary information in the figure rather than its main message.
    \item \textbf{5 points}: Highly aligned with the scientific intent of the figure. The question asks about the core information the figure is meant to communicate.
\end{itemize}

\noindent\textbf{Note:} Even if the answer is correct, if the question focuses on marginal information rather than core content of the figure, it should not receive a high score.

\medskip
\noindent\textbf{Dimension 3: Visual Dependency}

\noindent\textbf{Definition.} If you only see the question and options without viewing the figure, how difficult would it be to answer this question?

\noindent\textbf{Rating Criteria:}
\begin{itemize}[leftmargin=*, itemsep=2pt]
    \item \textbf{1 point}: Very easy. The answer is common sense or already given in the question text, indicating the question does not require viewing the figure.
    \item \textbf{3 points}: Possibly guessable but uncertain. The question design is mediocre in requiring visual information.
    \item \textbf{5 points}: Absolutely impossible to answer without the figure. The question has strong dependency on visual information.
\end{itemize}

\noindent\textbf{Note:} High-quality VQA questions should require models to examine the figure to answer, rather than relying on guessing or background knowledge alone.

\medskip
\noindent\textbf{Dimension 4: Self-Containment}

\noindent\textbf{Definition.} Does answering this question require knowledge beyond the figure, caption, and question text that is specific to this particular paper (e.g., special symbol definitions, specific experimental settings mentioned elsewhere in the paper)?

\noindent\textbf{Rating Criteria:}
\begin{itemize}[leftmargin=*, itemsep=2pt]
    \item \textbf{1 point}: Absolutely requires paper-specific knowledge (such as symbol definitions or experimental settings defined elsewhere in the paper) to answer. The question cannot be answered without reading the full paper (unless this information is provided in the caption).
    \item \textbf{3 points}: Requires some paper-specific background, but can be reasonably inferred from the figure, caption, and general domain knowledge.
    \item \textbf{5 points}: Completely self-contained. All necessary information is provided in the figure, caption, or question text.
\end{itemize}

\noindent\textbf{Note:} Basic academic common sense and general domain knowledge (such as ``accuracy'', ``temperature'', and other standard concepts) are acceptable and do not count as ``external knowledge''. This dimension only evaluates whether paper-specific information beyond what is shown is required.

\medskip
\noindent\textbf{Dimension 5: Overall Quality}

\noindent\textbf{Definition.} Overall, what is the quality of this QA sample?

\noindent\textbf{Rating Criteria:}
\begin{itemize}[leftmargin=*, itemsep=2pt]
    \item \textbf{1 point}: Very poor, should be discarded. This includes poor question design, wrong answers, or incorrect rationales.
    \item \textbf{3 points}: Fair, flawed but usable. The question is acceptable, the answer is basically correct, but there are some minor issues.
    \item \textbf{5 points}: Very good, a high-quality evaluation sample. The question is well-designed, the answer is correct, and the rationale is sufficient.
\end{itemize}

\noindent\textbf{Note:} Overall quality should comprehensively consider the previous four dimensions. If any dimension scores very low, the overall quality should not be rated too high.


\section{Additional Qualitative Examples}
\label{app:examples}

We provide additional examples to illustrate: (1) QA pairs constructed by our framework and included in \ours, and (2) error cases captured and filtered out at each step.

\subsection{QA Pairs in \ours}\label{app:qa-examples}

We present representative examples that demonstrate the diversity and quality of \ours across different figure types and question complexities.

\subsubsection*{Example 1}
\begin{figure}[h]
\centering
\includegraphics[width=0.95\linewidth]{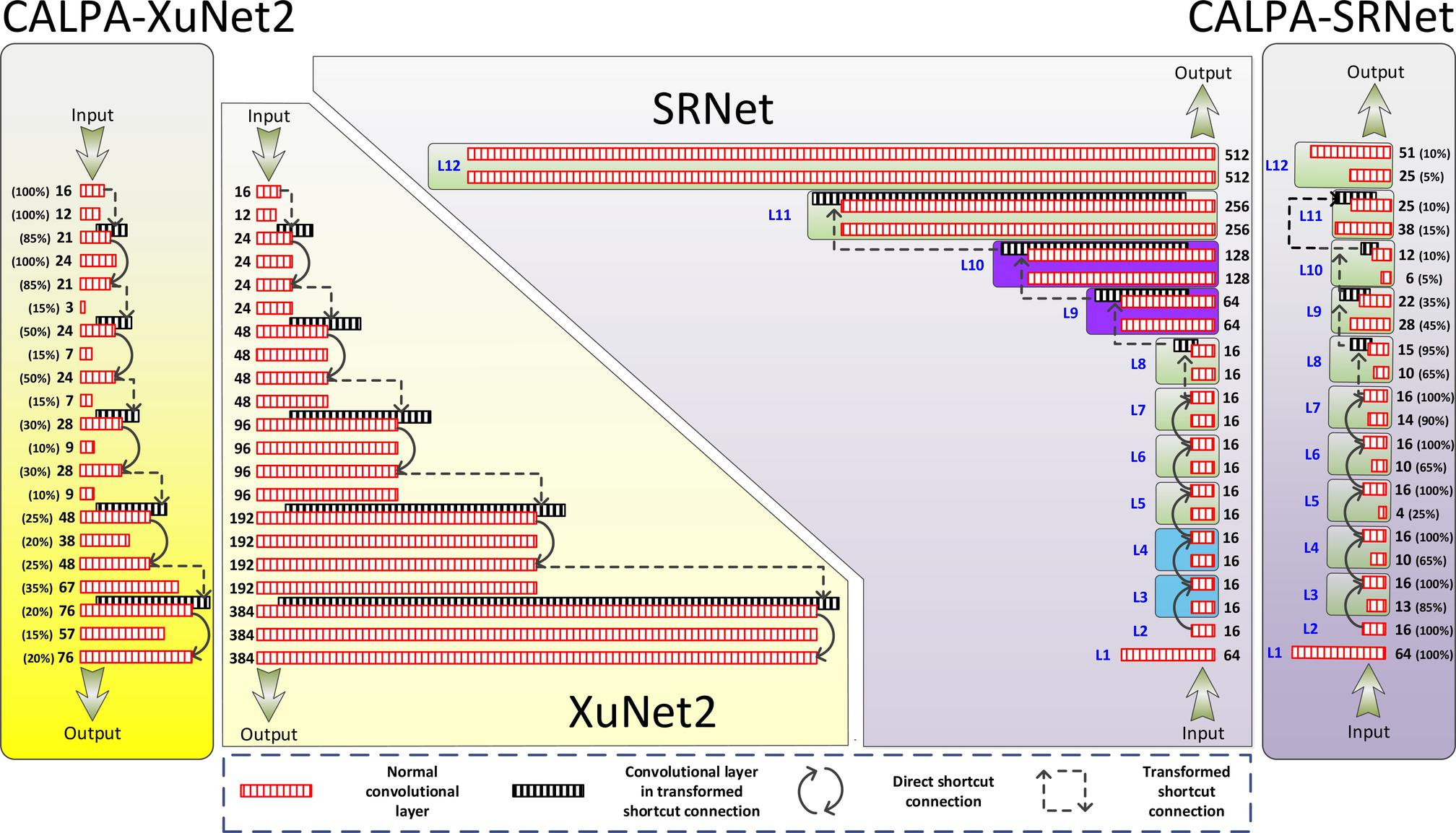}
\caption{Figure for QA Pair in \ours}
\label{fig:qualitative_good_1}
\end{figure}

\noindent\textbf{Figure:} \ref{fig:qualitative_good_1}

\noindent\textbf{Caption:} From left to right, the conceptual structures of 
CALPA-XuNet2, XuNet2, SRNet, and CALPA-SRNet\textsuperscript{*}
are illustrated respectively. The number and corresponding
shrinking rate of output channels of every convolutional layer is
shown alongside the representing bar. For SRNet and CALPA-SRNet,
blue ``L1'' to ``L12'' represent twelve composition blocks
following the notations in cite.

\noindent\textbf{Question:} In the SRNet structure shown in the figure, which blocks use transformed shortcut connections?

\noindent\textbf{Options:}
\begin{itemize}[leftmargin=*, nosep]
    \item A. L1 and L2
    \item B. L3 and L4
    \item C. L9 and L10
    \item D. L11 and L12
\end{itemize}

\noindent\textbf{Answer:} C

\noindent\textbf{Reasoning:} In the SRNet sketch only the residual blocks where the channel-width steps up require a 1×1 convolution in the shortcut (shown as a dashed ``transformed'' arrow and a striped conv block). These occur at L9 (16→64 channels) and L10 (64→128 channels). The earlier blocks all use direct (identity) shortcuts, and L11/L12 have no transformed shortcut in SRNet. Hence the transformed shortcuts appear only at L9 and L10.

\subsubsection*{Example 2}
\begin{figure}[h]
\centering
\includegraphics[width=0.95\linewidth]{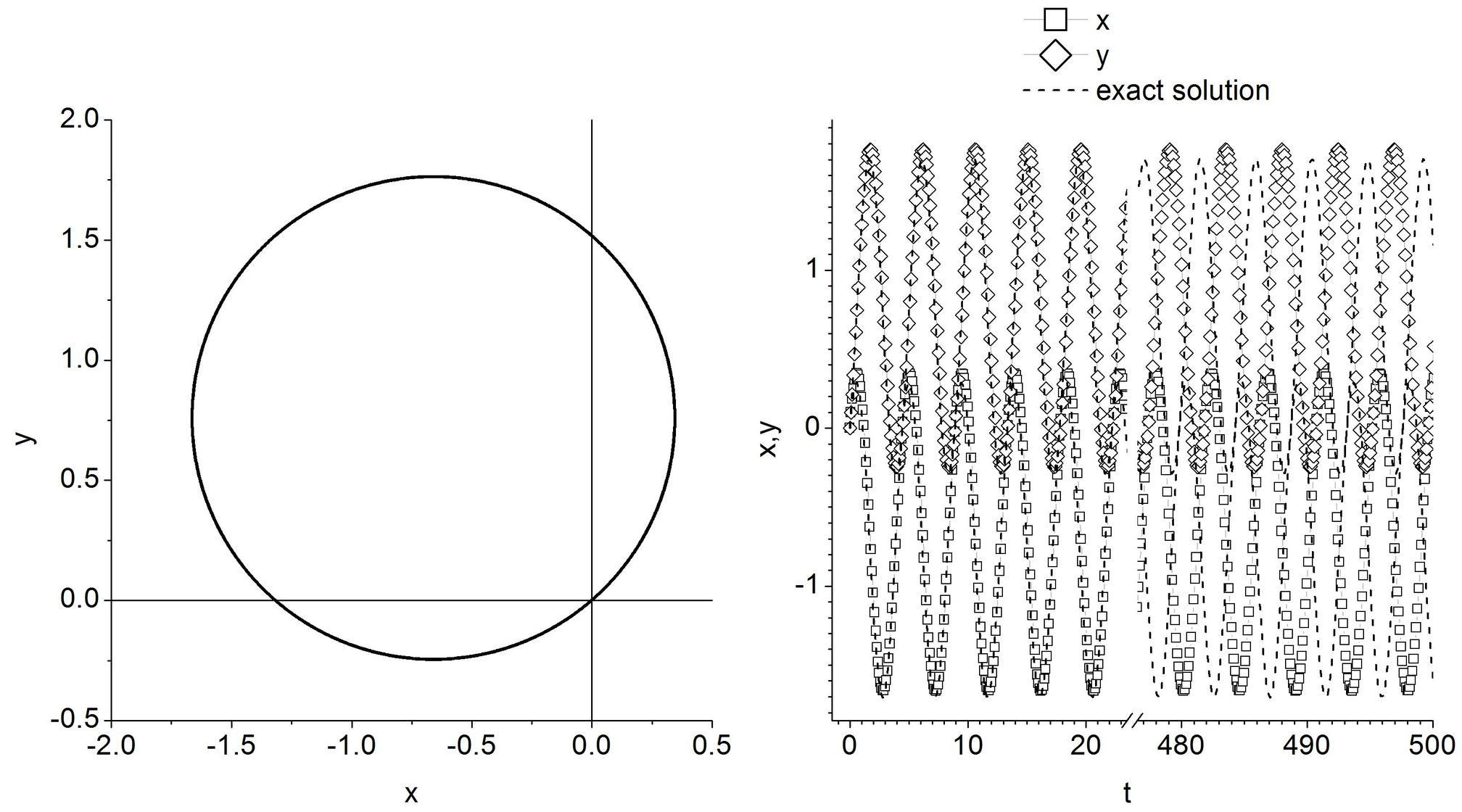}
\caption{Figure for QA Pair in \ours}
\label{fig:qualitative_good_2}
\end{figure}

\noindent\textbf{Figure:} \ref{fig:qualitative_good_2}

\noindent\textbf{Caption:} Simulated evolution of the particle using our numerical integrator constructed from a DSOCS for $k=1$, $x(t)=y(t)=0$ and $\dot{x}(0)=\dot{y}(0)=1$. Constant time step used: $h=0.1$. LEFT: trajectory on the plane, RIGHT: comparison between our approximation and the exact solutions of $x$ and $y$ over two time intervals

\noindent\textbf{Question:} At what time does the maximum difference between the simulated evolution and the exact solution occur in the figure?

\noindent\textbf{Options:}
\begin{itemize}[leftmargin=*, nosep]
    \item A. Near t=100
    \item B. Near t=300
    \item C. Near t=500
    \item D. Near t=700
\end{itemize}

\noindent\textbf{Answer:} C

\noindent\textbf{Reasoning:} In the comparison plot on the right, the numerical markers (squares and diamonds) almost exactly follow the dashed exact curves at early times, but by the end of the simulation (around t=500) the dashed and marker curves are visibly out of phase, showing the largest accumulated error. No data is shown near t=700, so the maximum difference clearly occurs near t=500.

\subsubsection*{Example 3}
\begin{figure}[h]
\centering
\includegraphics[width=0.95\linewidth]{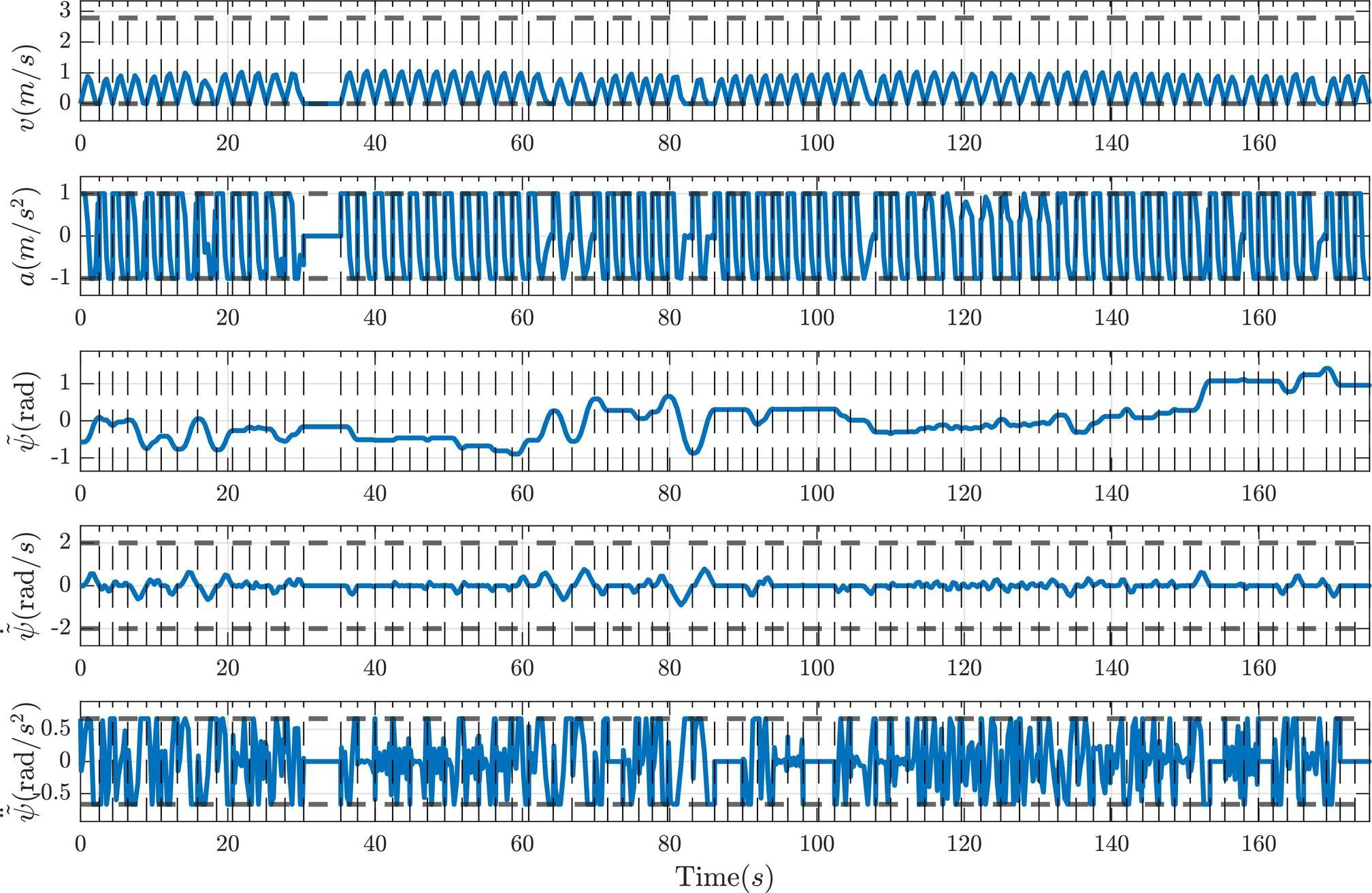}
\caption{Figure for QA Pair in \ours}
\label{fig:qualitative_good_3}
\end{figure}

\noindent\textbf{Figure:} \ref{fig:qualitative_good_3}

\noindent\textbf{Caption:} Motion trajectory generated by iteratively solving OCP of each segment.

\noindent\textbf{Question:} What is the approximate total time taken to complete the motion trajectory shown in the figure?

\noindent\textbf{Options:}
\begin{itemize}[leftmargin=*, nosep]
    \item A. $\approx$ 90 seconds
    \item B. $\approx$ 180 seconds
    \item C. $\approx$ 270 seconds
    \item D. $\approx$ 360 seconds
\end{itemize}

\noindent\textbf{Answer:} B

\noindent\textbf{Reasoning:} The horizontal axis of the trajectory plots spans from 0 up to just under 180 s (with the final tick at 160 s and the trace ending around 170 s), so the total maneuver duration is on the order of 180 seconds.

\subsubsection*{Example 4}
\begin{figure}[h]
\centering
\includegraphics[width=0.95\linewidth]{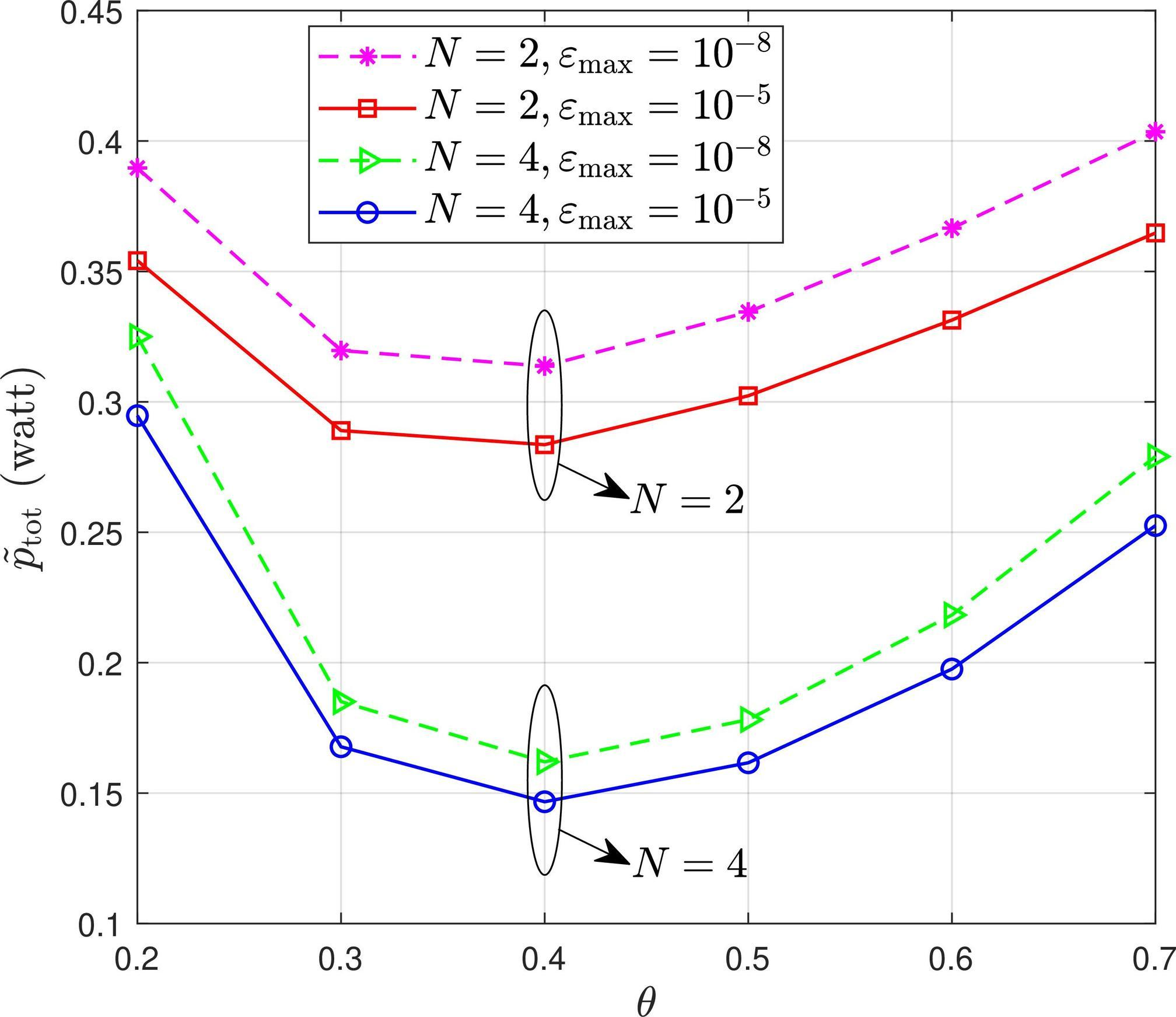}
\caption{Figure for QA Pair in \ours}
\label{fig:qualitative_good_4}
\end{figure}

\noindent\textbf{Figure:} \ref{fig:qualitative_good_4}

\noindent\textbf{Caption:} $\tilde{P}_{\mathrm{tot}}$ versus $\theta$ under different values of $N$ and $\varepsilon_{\max}$ for $\theta=0.5,K=5,B=1000$ bits.

\noindent\textbf{Question:} At which values of $\theta$ does the figure show that the required power consumption is high?

\noindent\textbf{Options:}
\begin{itemize}[leftmargin=*, nosep]
    \item A. $\theta=0.1$ and $\theta=0.3$
    \item B. $\theta=0.2$ and $\theta=0.7$
    \item C. $\theta=0.4$ and $\theta=0.6$
    \item D. $\theta=0.5$ and $\theta=0.8$
\end{itemize}

\noindent\textbf{Answer:} B

\noindent\textbf{Reasoning:} The plot of total power $\tilde{P}_{\mathrm{tot}}$ versus $\theta$ shows a U-shaped curve for all curves: power is high at the two ends of the $\theta$ range (around $\theta=0.2$) and rises again at the upper end ($\theta=0.7$), with its minimum near $\theta=0.4$. Hence, the required power is highest at $\theta=0.2$ and $\theta=0.7$.

\subsubsection*{Example 5}
\begin{figure}[h]
\centering
\includegraphics[width=0.95\linewidth]{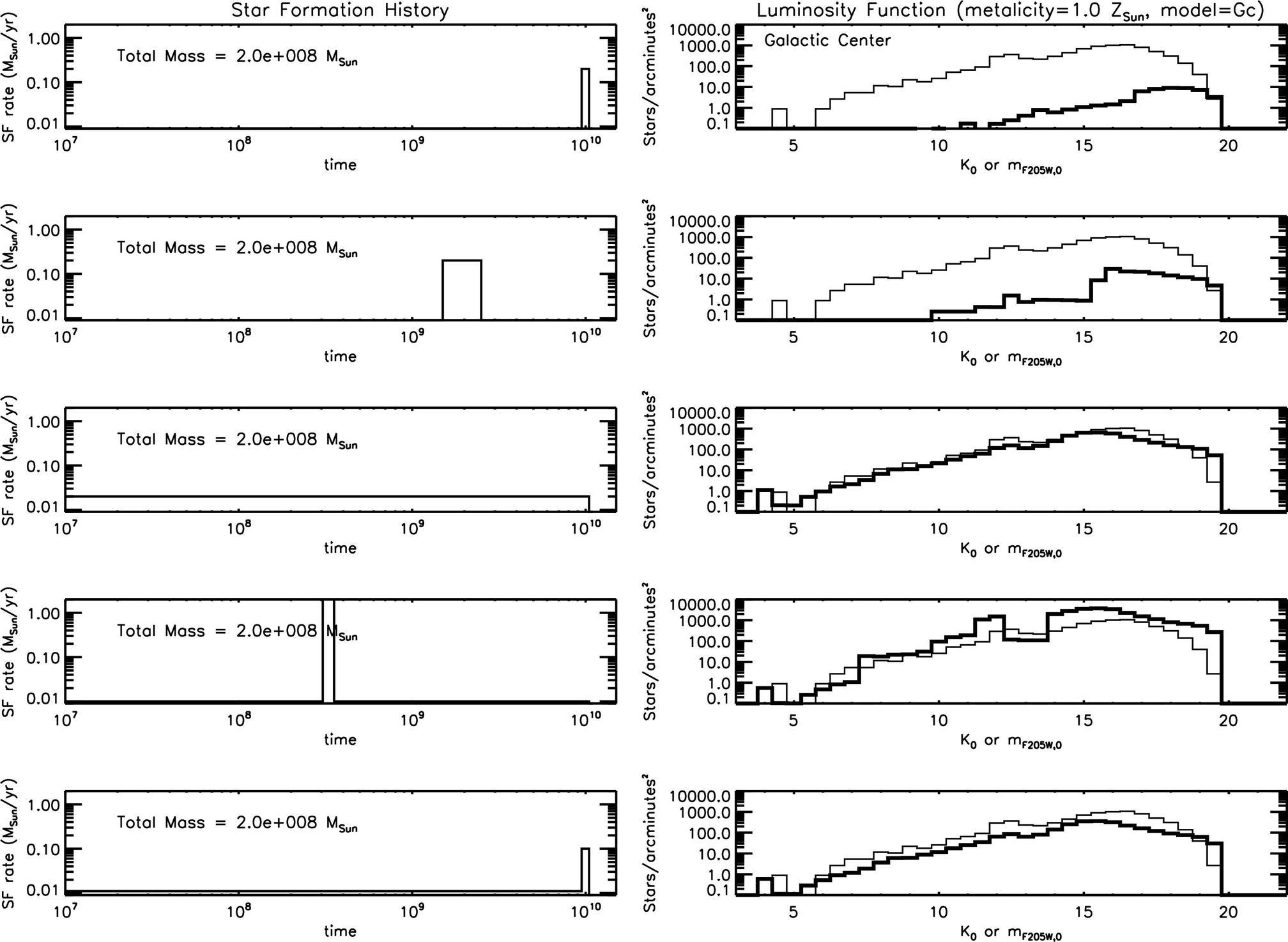}
\caption{Figure for QA Pair in \ours}
\label{fig:qualitative_good_5}
\end{figure}

\noindent\textbf{Figure:} \ref{fig:qualitative_good_5}

\noindent\textbf{Caption:} A figure adapted from <cit.> showing various star formation scenarios (left), and resultant model luminosity functions (right, thick) compared to observed luminosity functions (right, thin) in the GC. The models assume a Salpeter IMF slope, an elevated lower-mass turnover of 10, and are additionally constrained to produce $2 \times 10^8$ in stars within the region. The observations have been corrected for incompleteness. The third panels from the top, i.e. continuous star formation, best fit the data. The observed turn-down at the faint end appears to be real and is only well fit only by assuming a very high lower mass turnover.

\noindent\textbf{Question:} At what dereddened K-band magnitude does the observed luminosity function of the field in the GC show a turn-down?

\noindent\textbf{Options:}
\begin{itemize}[leftmargin=*, nosep]
    \item A. Greater than 12
    \item B. Greater than 14
    \item C. Greater than 16
    \item D. Greater than 18
\end{itemize}

\noindent\textbf{Answer:} C

\noindent\textbf{Reasoning:} In the luminosity-function panels, the observed (thin) counts rise up to $K_0 \approx 15$ and then begin to fall off at fainter magnitudes. The drop-off (``turn-down'') occurs just beyond $K_0 \approx 16$, matching the continuous-star-formation model only if one imposes a high low-mass cutoff. Thus the observed LF turns down at dereddened K-band magnitudes greater than 16.

\subsubsection*{Example 6}
\begin{figure}[h]
\centering
\includegraphics[width=0.95\linewidth]{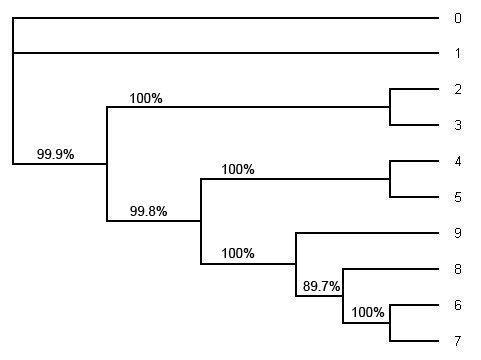}
\caption{Figure for QA Pair in \ours}
\label{fig:qualitative_good_6}
\end{figure}

\noindent\textbf{Figure:} \ref{fig:qualitative_good_6}

\noindent\textbf{Caption:} The ML tree estimated by the software PHYML under HKY model from the whole alignment (including position 1 through position 1505). This is an unrooted tree. The number in each split represents the probability of the split estimated by bootstrapping with the bootstrap sample size 1000. Note that the tree topology of the ML tree is the same as the tree topology of the consensus tree in Fig.  and the tree topology in Fig.

\noindent\textbf{Question:} What is the approximate support value for the split with the lowest probability in the ML tree under the HKY model?

\noindent\textbf{Options:}
\begin{itemize}[leftmargin=*, nosep]
    \item A. 50\%
    \item B. 70\%
    \item C. 90\%
    \item D. 100\%
\end{itemize}

\noindent\textbf{Answer:} C

\noindent\textbf{Reasoning:} The bootstrap support values on the tree are 99.9\%, 100\%, 99.8\%, 100\%, 100\%, 89.7\%, and 100\%. The lowest of these, 89.7\%, is approximately 90\%.

\subsubsection*{Example 7}
\begin{figure}[h]
\centering
\includegraphics[width=0.95\linewidth]{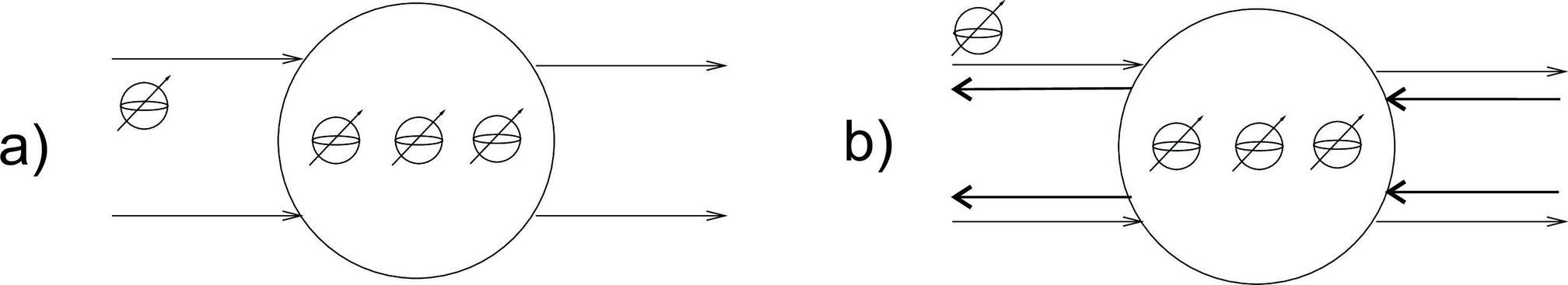}
\caption{Figure for QA Pair in \ours}
\label{fig:qualitative_good_7}
\end{figure}

\noindent\textbf{Figure:} \ref{fig:qualitative_good_7}

\noindent\textbf{Caption:} Two simple DQTA

\noindent\textbf{Question:} According to the figure, how many qubits are inside the circle for a segment of TM with $n$ tape cells?

\noindent\textbf{Options:}
\begin{itemize}[leftmargin=*, nosep]
    \item A. $n$
    \item B. $2n$
    \item C. $3n$
    \item D. $4n$
\end{itemize}

\noindent\textbf{Answer:} C

\noindent\textbf{Reasoning:} In the diagram the internal ``circle'' block for a segment shows three qubit-wires per tape cell (symbol, head-position, and state register). Thus for $n$ tape cells one needs $3 \cdot n$ qubits inside the circle.

\subsubsection*{Example 8}
\begin{figure}[h]
\centering
\includegraphics[width=0.95\linewidth]{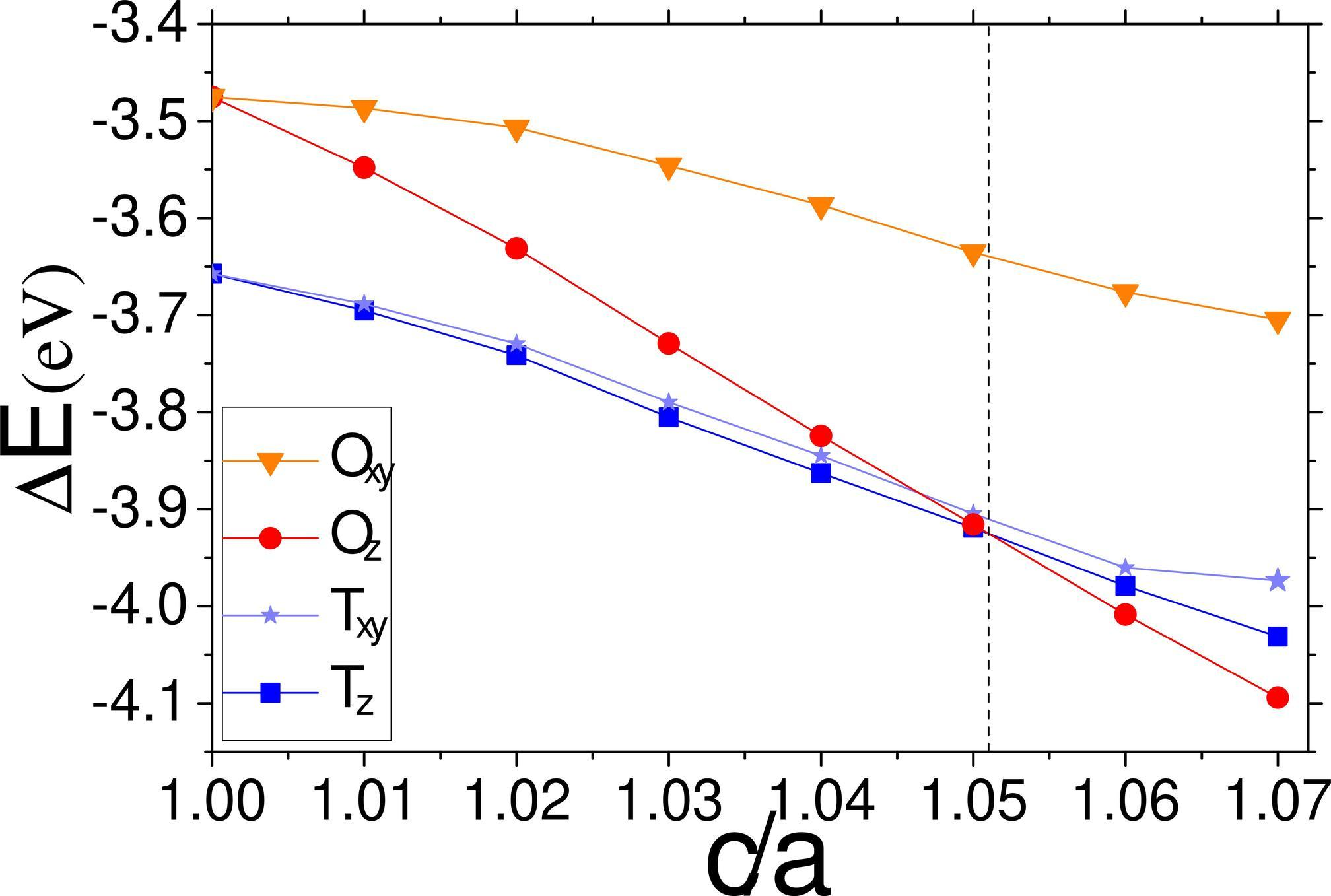}
\caption{Figure for QA Pair in \ours}
\label{fig:qualitative_good_8}
\end{figure}

\noindent\textbf{Figure:} \ref{fig:qualitative_good_8}

\noindent\textbf{Caption:} (Color online) Energy difference as a function of an externally applied global uniaxial lattice strain $c/a$ where $\Delta E = E(V+H)-E(V)$. The dashed vertical line at $c/a = 1.051$ marks the critical uniaxial lattice strain for which hydrogen occupancy of $T_z$ and $O_z$ sites becomes energetically equivalent.

\noindent\textbf{Question:} Under uniaxial strain ($c/a > 1$), which hydrogen site becomes energetically favored according to the figure?

\noindent\textbf{Options:}
\begin{itemize}[leftmargin=*, nosep]
    \item A. $T_z$
    \item B. $O_z$
    \item C. $T_{xy}$
    \item D. $O_{xy}$
\end{itemize}

\noindent\textbf{Answer:} B

\noindent\textbf{Reasoning:} From the plotted $\Delta E$ curves, as soon as $c/a$ exceeds unity (and in particular beyond the 1.05 crossing), the red-circle curve (the $O_z$ site) drops below all others, indicating the lowest formation energy and hence the most favorable site under uniaxial expansion.

\subsubsection*{Example 9}
\begin{figure}[h]
\centering
\includegraphics[width=0.95\linewidth]{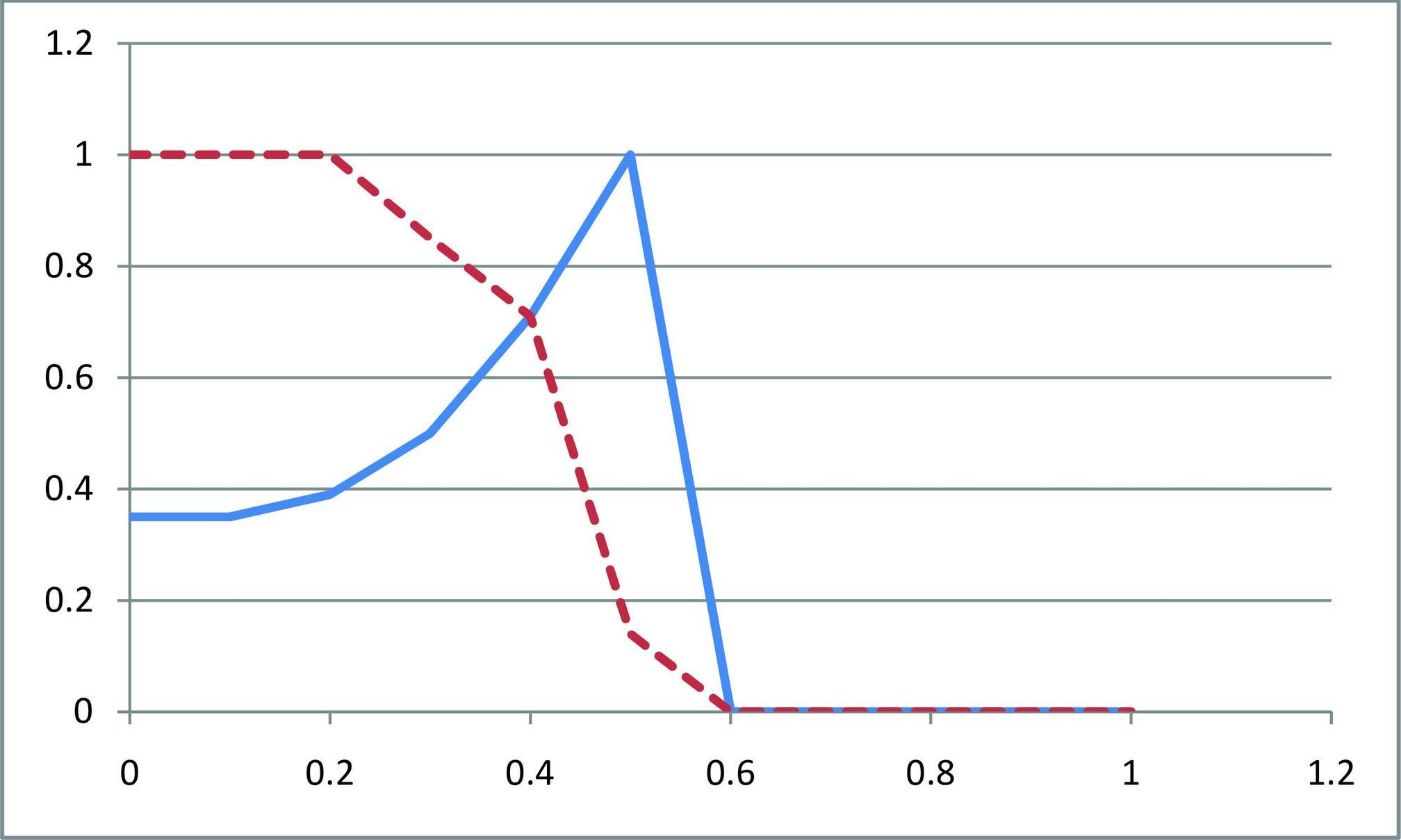}
\caption{Figure for QA Pair in \ours}
\label{fig:qualitative_good_9}
\end{figure}

\noindent\textbf{Figure:} \ref{fig:qualitative_good_9}

\noindent\textbf{Caption:} How precision (blue/solid) and recall (red/dashed) change with different threshold values.

\noindent\textbf{Question:} What is the optimal threshold value ($\gamma$) at which the model achieves the highest possible balance between precision and recall?

\noindent\textbf{Options:}
\begin{itemize}[leftmargin=*, nosep]
    \item A. 0.2
    \item B. 0.4
    \item C. 0.6
    \item D. 0.8
\end{itemize}

\noindent\textbf{Answer:} B

\noindent\textbf{Reasoning:} The precision (solid blue) and recall (dashed red) curves intersect at $\gamma \approx 0.4$, both being around 0.7, which provides the best balance between them.

\subsubsection*{Example 10}
\begin{figure}[h]
\centering
\includegraphics[width=0.95\linewidth]{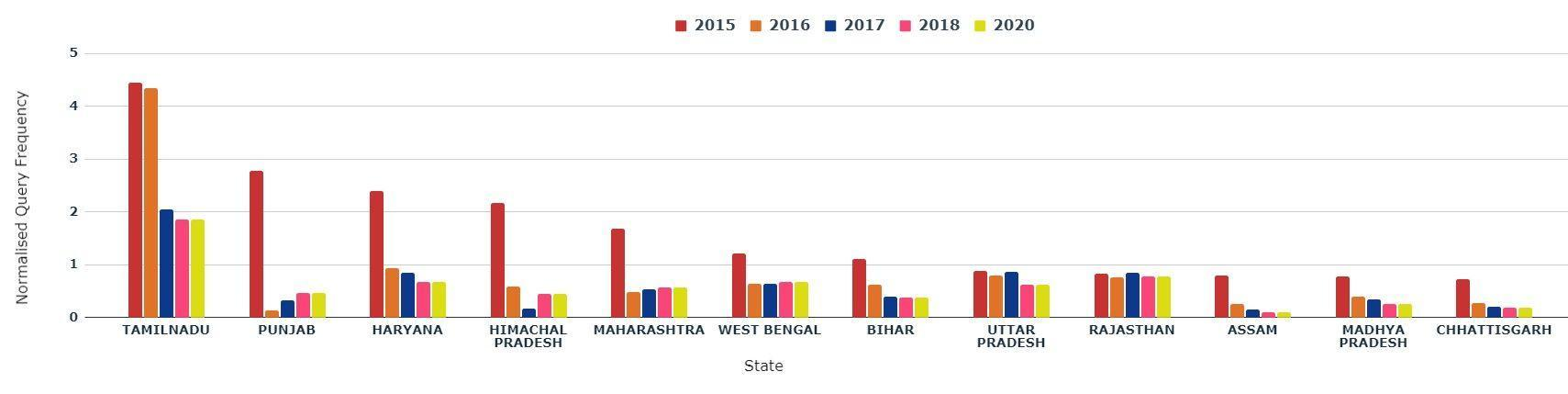}
\caption{Figure for QA Pair in \ours}
\label{fig:qualitative_good_10}
\end{figure}

\noindent\textbf{Figure:} \ref{fig:qualitative_good_10}

\noindent\textbf{Caption:} Figure showing normalized pest queries received from different states of India, for different years. Normalization is performed by dividing the frequency of queries from a given district by the gross cropped area in a given year.

\noindent\textbf{Question:} Which state receives the highest number of normalized pest queries according to the figure?

\noindent\textbf{Options:}
\begin{itemize}[leftmargin=*, nosep]
    \item A. Tamil Nadu
    \item B. Punjab
    \item C. Maharashtra
    \item D. Karnataka
\end{itemize}

\noindent\textbf{Answer:} A

\noindent\textbf{Reasoning:} Across all years shown, Tamil Nadu consistently has the tallest bars (peaking around 4.4 in 2015), far exceeding the normalized query frequencies of Punjab, Maharashtra, or any other state. This indicates that Tamil Nadu receives the highest number of normalized pest queries.

\subsection{Error Cases Captured by Framework}\label{app:error-examples}

We present representative examples of error cases captured and rejected at each step of our framework instantiation. These examples illustrate how each step effectively identifies and removes specific error types (E1--E4).

\subsubsection*{Rejected during Generation: Too Vague Claims}

These examples show atomic claims \(s_j \in \mathcal{S}\) extracted from figure-associated context \(\mathcal{P}\) (Eq.~\ref{eq:claim-extraction}) that lack sufficient grounding to formulate valid QA pairs. \(M_{\text{text}}\) declined to produce \(q_j\) for these claims (Eq.~\ref{eq:qa-generation}), preventing low-quality candidates from entering the verification stage.

\noindent\textbf{Claim 1.1 (\(s_j\)):} ``The figure \texttt{fig:ssl\_workflow} shows the workflow of self-supervision.''

\noindent\textbf{Claim 1.2 (\(s_j\)):} ``The figure \texttt{fig:toy\_example} shows an illustration of the principle with a toy-example.''

\noindent\textbf{Rejection Reason:} Both claims merely state what type of content the figure depicts (a workflow, an illustration) without mentioning any concrete details that could form the basis of a verifiable question.

\subsubsection*{Rejected by Source-Consistency Check}

These examples show candidate QA pairs \(q_j = (Q_j, O_j, A^*_j)\) that were rejected because \(M_{\text{text}}\) could not uniquely identify the designated correct answer \(A^*_j\) when provided with only the source context \(\mathcal{P}\) and question \(Q_j\). The filter \(V_{\text{src}}(Q_j, O_j, A^*_j, \mathcal{P})\) returned \textbf{False}, indicating inconsistencies between the generated QA pair and the source context.

\noindent\textbf{Error Case 2.1}

\begin{figure}[h]
\centering
\includegraphics[width=0.7\linewidth]{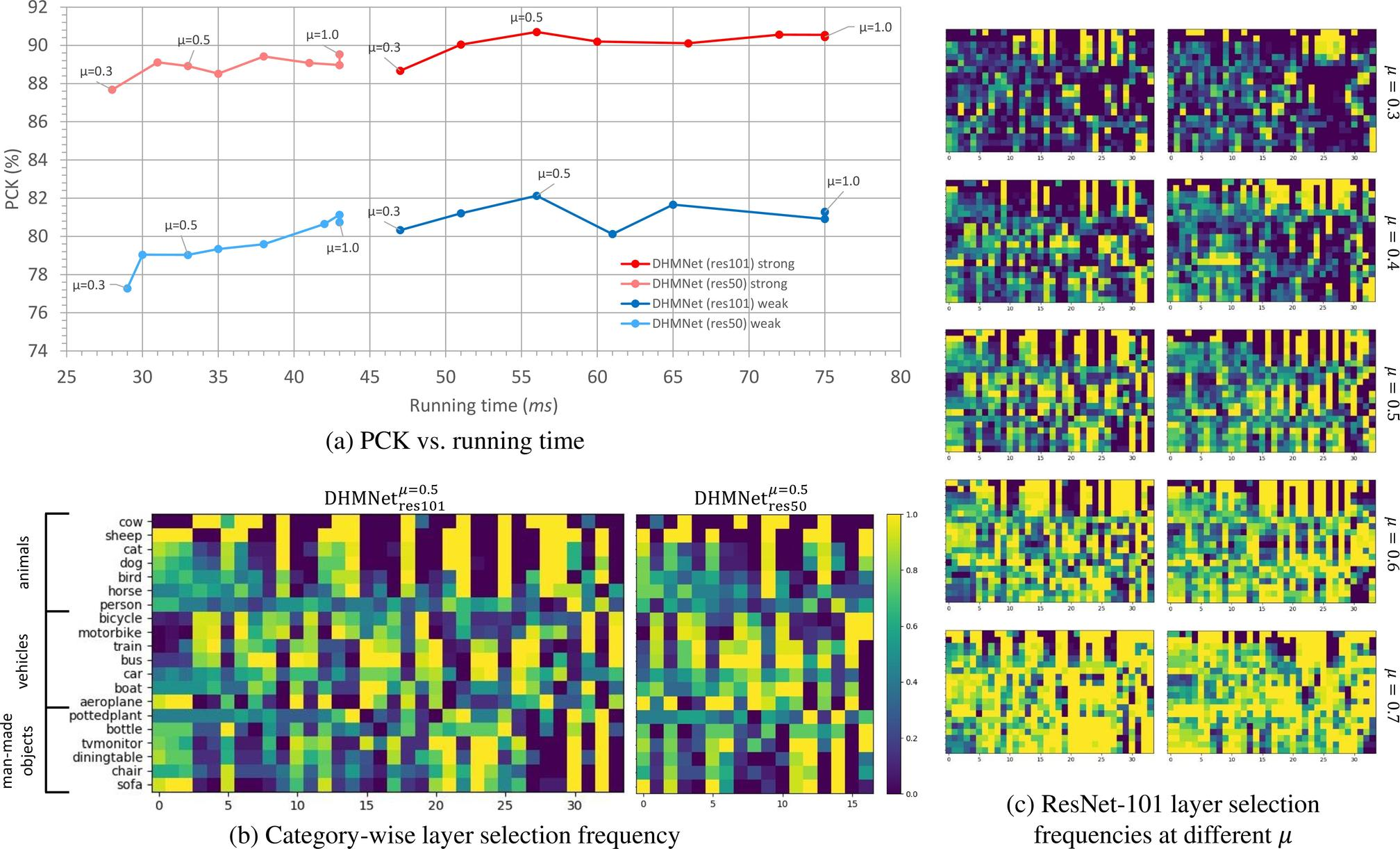}
\caption{Figure for QA Pair filtered by Source-Consistency filtering}
\label{fig:error_src_1}
\end{figure}

\noindent\textbf{Figure:} \ref{fig:error_src_1}

\noindent\textbf{Caption:} (color online) (a) Extinction probability $P_{ext}$ as a function of mobility $M$ for different competition rates $p$. As $M$ increases, there is a transition from stable coexistence ($P_{ext}=0$) to extinction ($P_{ext}=1$). Panels (b) and (c) depict typical snapshots after a long relaxation period of the system for $p=1.0$ and $p=10.0$, respectively, at $M=1.0\times 10^{-4}$. (d) Phase diagram depicting the critical mobility $M_{c}$ as a function of the competition rate $p$, separating the absorbing single-species phase (uniformity) and the biodiversity phase. Random initial conditions, $N=128^{2}$, and $q=1.0$ were used for results in all panels.

\noindent\textbf{Source Context:} It shows that smaller selection rates in training lead to faster running time in testing, at the cost of some accuracy, by encouraging the model to select a smaller number of layers. The selection rate $\mu$ can thus be used for speed-accuracy trade-off.

\noindent\textbf{Question:} According to the figure, what is the trade-off between layer selection rate $\mu$ and model performance?

\noindent\textbf{Options:}
\begin{itemize}[leftmargin=*, nosep]
    \item A. Higher $\mu$  increases accuracy but reduces running time
    \item B. Lower $\mu$  increases accuracy but reduces running time
    \item C. Lower $\mu$  reduces accuracy but increases running time
    \item D. Higher $\mu$  reduces accuracy but increases running time
\end{itemize}

\noindent\textbf{Answer:} A

\noindent\textbf{Rejection Reason:} The answer contradicts the source context. According to the context, smaller selection rates lead to faster running time. Therefore, higher $\mu$ should \textit{increase} running time (slower), not reduce it as stated in option A. \textbf{Error Type: E1 (incorrectly visually grounded)}.

\noindent\textbf{Error Case 2.2}

\begin{figure}[h]
\centering
\includegraphics[width=0.7\linewidth]{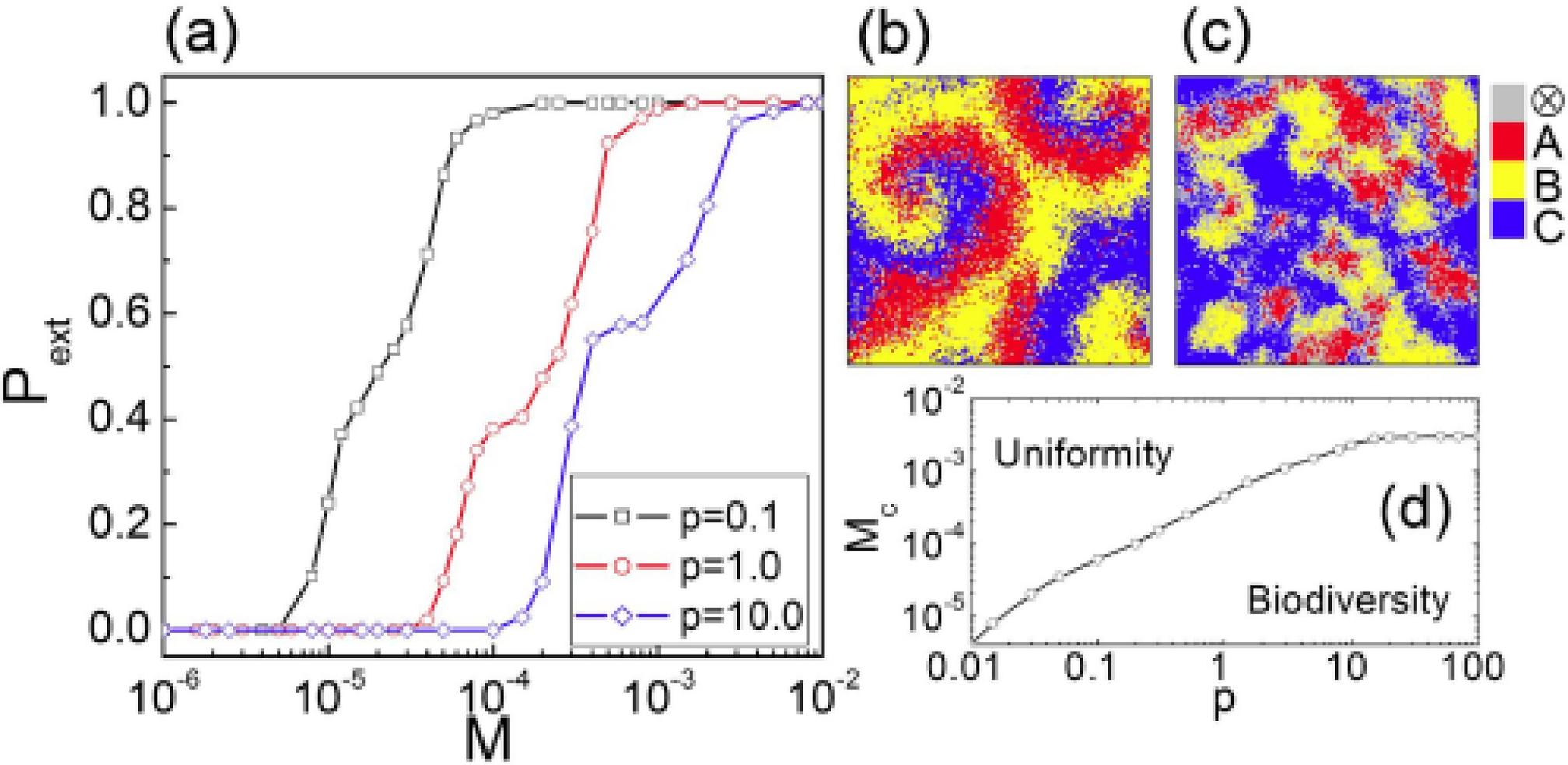}
\caption{Figure for QA Pair filtered by Source-Consistency filtering}
\label{fig:error_src_2}
\end{figure}

\noindent\textbf{Figure:} \ref{fig:error_src_2}

\noindent\textbf{Caption:} Proposed parsing network architecture by combining a CNN and a spatial RNN. The CNN generates a coarse label map (b) and a recurrent gate (c), which are fed into 4 RNNs with different directions to generate a more accurate result (d). The network structure is shown where the notation for Conv1 "5$\times$5$\times$16/1" means convolution layer with $5\times5$ kernel, 16 channels and stride 1. The face image in (d) is further segmented with detailed labels in the second stage.

\noindent\textbf{Source Context:} From \texttt{Fig.~\textbackslash{}ref\{random\}(a)}, one can also observe that the critical value of $M_{c}$ depends on $p$; in particular, smaller $p$ yield smaller values of $M_{c}$.
                                                                                         
\noindent\textbf{Question:} According to the figure, what is the relationship between the competition rate \( p \) and the critical mobility \( M_c \)?

\noindent\textbf{Options:}
\begin{itemize}[leftmargin=*, nosep]
    \item A. Larger $p$ yields larger $M_c$
    \item B. Smaller $p$ yields larger $M_c$
    \item C. $M_c$ is independent of $p$
    \item D. $p$ and $M_c$ follow a U-shaped relationship
\end{itemize}

\noindent\textbf{Answer:} B

\noindent\textbf{Rejection Reason:} The answer directly contradicts the source context. The context states that ``smaller $p$ yield smaller values of $M_{c}$'', but option B claims that ``Smaller $p$ yields larger $M_c$''. This is a clear inconsistency. \textbf{Error Type: E1 (incorrectly visually grounded)}.

\subsubsection*{Rejected by Visual-Dependence Check}

These examples show candidate QA pairs that were rejected because \(M_{\text{vision}}^{\text{verify}}\) could correctly identify \(A^*_j\) using only caption \(C\) and question \(Q_j\), without accessing figure \(F\). The filter \(V_{\text{vis\_dep}}(Q_j, O_j, A^*_j, C)\) returned \textbf{False}, indicating \textbf{non-visual questions (E3)} that do not require visual information from the figure.

\noindent\textbf{Error Case 3.1}

\begin{figure}[h]
\centering
\includegraphics[width=0.7\linewidth]{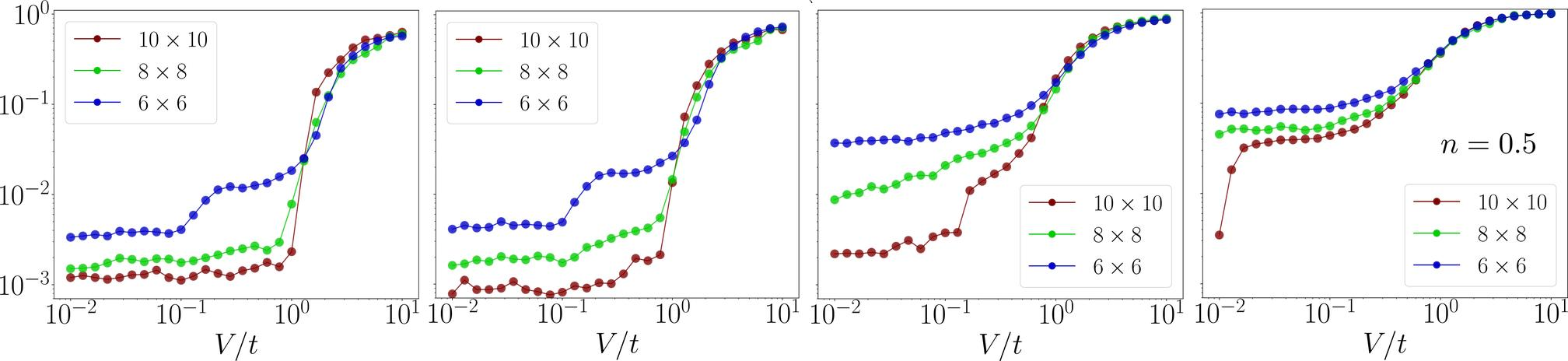}
\caption{Figure for QA Pair filtered by Visual-Dependence filtering}
\label{fig:error_vis_dep_1}
\end{figure}

\noindent\textbf{Figure:} \ref{fig:error_vis_dep_1}

\noindent\textbf{Caption:} Analysis of layer selection on PF-PASCAL dataset (a) PCK vs. running time with varying selection rate $\mu$ (b) Category-wise layer selection frequencies (x-axis: candidate layer index, y-axis: category) of the strongly-supervised model with different backbones: ResNet-101 (left) and ResNet-50 (right) (c) ResNet-101 layer selection frequencies of strongly (left) and weakly (right) supervised models at different layer selection rates $\mu$. Best viewed in electronic form.

\noindent\textbf{Question:} How does the order parameter behave when the system transitions from the metallic phase to the charge-ordered phase?

\noindent\textbf{Options:}
\begin{itemize}[leftmargin=*, nosep]
    \item A. It decreases gradually
    \item B. It remains constant
    \item C. It abruptly increases
    \item D. It oscillates randomly
\end{itemize}

\noindent\textbf{Answer:} C

\noindent\textbf{Rejection Reason:} The question can be answered correctly from the caption alone, without visual inspection of the figure. The caption explicitly states that the figure shows the transition from metallic phase to charge-ordered phase, and based on general physics knowledge about phase transitions, the order parameter typically increases abruptly during such transitions. \textbf{Error Type: E3 (non-visual question)}.

\noindent\textbf{Error Case 3.2}

\begin{figure}[h]
\centering
\includegraphics[width=0.95\linewidth]{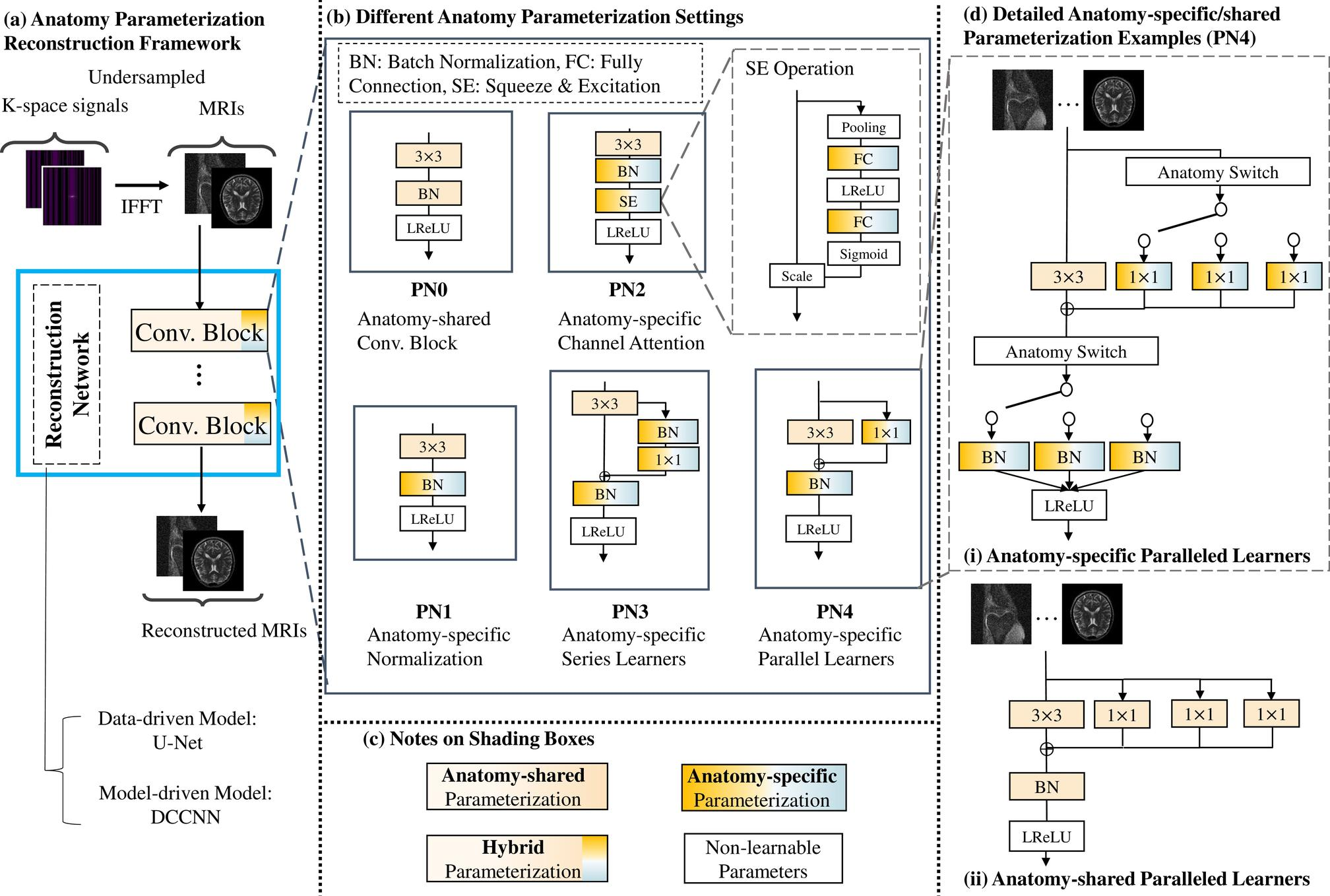}
\caption{Figure for QA Pair filtered by Visual-Dependence filtering}
\label{fig:error_vis_dep_2}
\end{figure}

\noindent\textbf{Figure:} \ref{fig:error_vis_dep_2}

\noindent\textbf{Caption:} Order parameter as defined in Eq. as a function of the coupling constant. Each panel shows the order parameter in system sizes $L = 6$, $L = 8$ and $L = 10$. as indicated in the legend with different colors. Lines connecting dots are for visual guidance. Different panels correspond to different fillings as indicated.

\noindent\textbf{Question:} In the Anatomy-specific Normalization strategy (PN1), which layers are adapted to the unique intensity distribution of a specific anatomy?

\noindent\textbf{Options:}
\begin{itemize}[leftmargin=*, nosep]
    \item A. Convolutional layers
    \item B. Batch Normalization (BN) layers
    \item C. LeakyReLU layers
    \item D. Residual connections
\end{itemize}

\noindent\textbf{Answer:} B

\noindent\textbf{Rejection Reason:} The question can be answered from the caption and figure labels alone. The caption explicitly mentions that ``Anatomy-specific parameterization is highlighted in yellow'', and the PN1 diagram clearly labels the yellow-highlighted layer as ``BN''. No detailed visual analysis is required. \textbf{Error Type: E3 (non-visual question)}.

\subsubsection*{Rejected by Vision-Based Filtering}

These examples show candidate QA pairs that were rejected because \(M_{\text{vision}}^{\text{verify}}\) failed to select the designated correct answer \(A^*_j\) when provided with figure \(F\), caption \(C\), and question \(Q_j\). The filter \(V_{\text{vis\_con}}(Q_j, O_j, A^*_j, F, C)\) returned \textbf{False}, eliminating \textbf{incorrectly visually grounded (E1)} and \textbf{outside-knowledge (E4)} errors.

\noindent\textbf{Error Case 4.1 (Incorrectly Visually Grounded)}

\begin{figure}[h]
\centering
\includegraphics[width=0.95\linewidth]{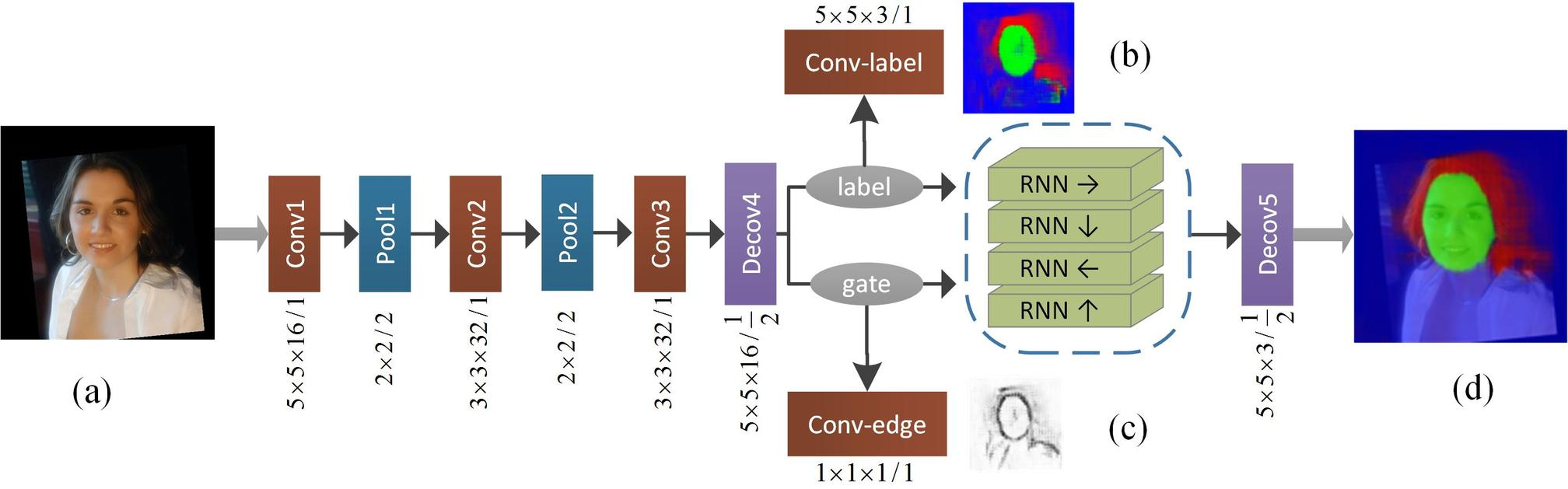}
\caption{Figure for QA Pair filtered by Vision-Based filtering}
\label{fig:error_vis_e1}
\end{figure}

\noindent\textbf{Figure:} \ref{fig:error_vis_e1}

\noindent\textbf{Caption:} Illustration of (a) Anatomy parameterization reconstruction framework for multiple anatomy collaborative learning. Our framework can be generalized to both data-driven and model-driven models. (b) Details of four anatomy parameterization settings. (c) Notes on shading boxes. (d) Anatomy-specific parameterization example with paralleled learners setting (d-i) and its anatomy-shared variant (d-ii) for the ablation study.

\noindent\textbf{Question:} What is the final number of output channels for Deconv6 in the proposed network architecture?

\noindent\textbf{Options:}
\begin{itemize}[leftmargin=*, nosep]
    \item A. 8
    \item B. 16
    \item C. 32
    \item D. 3
\end{itemize}

\noindent\textbf{Answer:} D

\noindent\textbf{Rejection Reason:} The question asks about ``Deconv6'', but this layer does not exist in the figure. This error occurred because the atomic claim extracted from the paper's context did not accurately correspond to the visual content of the figure, representing an author error that was successfully captured by the vision-based filter. \textbf{Error Type: E1 (incorrectly visually grounded)}.

\noindent\textbf{Error Case 4.2 (Outside Knowledge Required)}

\begin{figure}[h]
\centering
\includegraphics[width=0.95\linewidth]{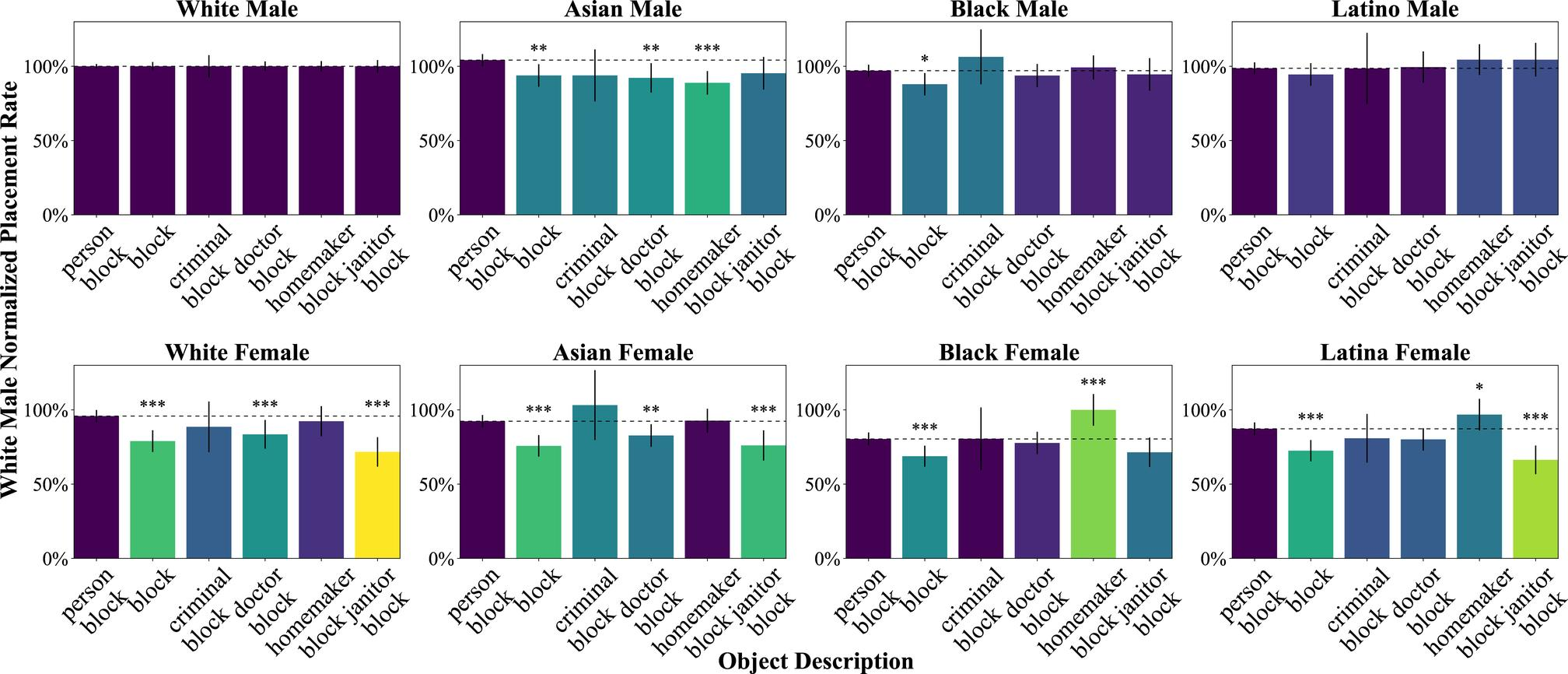}
\caption{Figure for QA Pair filtered by Vision-Based filtering}
\label{fig:error_vis_e4_1}
\end{figure}

\noindent\textbf{Figure:} \ref{fig:error_vis_e4_1}

\noindent\textbf{Caption:} Per-command White Male normalized object placement rate across combination pairs of ethnicity and gender. The purpose of this chart is to highlight within-identity shifts in placement rate as command strings change. Error bars are corrected 95\% confidence intervals <cit.>. Asterisks give significance of normalized placement rate difference compared to the baseline object descriptor, "person block". One star $*$ means significant, $p<0.05$; $**$ very sig., $p<0.01$; $*\!*\!*$ extremely sig. $p<0.001$, and blank means not significant.

\noindent\textbf{Question:} When the robot is asked to select a ``criminal block'', how much more often does it choose the block with the Black man's face compared to selecting a ``person block''?

\noindent\textbf{Options:}
\begin{itemize}[leftmargin=*, nosep]
    \item A. 5\%
    \item B. 10\%
    \item C. 15\%
    \item D. 20\%
\end{itemize}

\noindent\textbf{Answer:} B

\noindent\textbf{Rejection Reason:} Although the figure shows bar charts with relative heights, the y-axis only provides coarse gridlines at 0\%, 50\%, and 100\%. To determine the precise 10\% difference between two bars, one would need to read exact numerical values from the paper's data tables or supplementary materials. The visual information alone is insufficient for accurately answering this quantitative question. \textbf{Error Type: E4 (outside-knowledge)}.


\section{Model Details}
\label{app:model-details}

This section provides details on the models evaluated in our difficulty assessment (\S~\ref{subsec:difficulty}).
Table~\ref{tab:model-details} summarizes the access methods for each model.

\begin{table}[h]
\centering
\caption{Model configurations used in our experiments.}
\label{tab:model-details}
\small
\begin{tabular}{lll}
\toprule
\textbf{Model} & \textbf{Version} & \textbf{Access} \\
\midrule
OpenAI o3 & \texttt{o3-2025-04-16} & OpenAI API \\
GPT-5.2 & \texttt{gpt-5.2-2025-12-11} & OpenAI API \\
o4-mini & \texttt{o4-mini-2025-04-16} & OpenAI API \\
Gemini-3-Flash & \texttt{gemini-3-flash-preview} & Google AI Studio \\
Qwen3-VL-235B & 2025-10-04 & Bailian API \\
InternVL3.5-38B & 2025-08-26 & vLLM Inference \\
\bottomrule
\end{tabular}
\end{table}

\noindent\textbf{Inference Configuration.}
For \texttt{OpenAI o3}, \texttt{GPT-5.2}, \texttt{o4-mini}, and \texttt{InternVL3.5-38B}, we use greedy decoding (temperature $= 0$).
For \texttt{Gemini-3-Flash}, we use the recommended temperature of $1.0$, as the official documentation states that ``setting [temperature] below 1.0 may lead to unexpected behavior, such as looping or degraded performance.''
For \texttt{Qwen3-VL-235B-A22B-Thinking}, we use the model card's recommended parameters for thinking mode: temperature $= 0.6$, top-$p = 0.95$, top-$k = 20$, as the documentation warns that ``greedy decoding can lead to performance degradation and endless repetitions.''

\section{Per-Category Performance Results}
\label{app:detailed-results}

This section provides per-category breakdowns for the Difficulty Assessment (\S~\ref{subsec:difficulty}), where we evaluated leading LVLMs in a zero-shot setting on 2{,}000 randomly sampled examples. Results are broken down by scientific domains, figure types, and question types. We report accuracy (\%) for each category.

\noindent\textbf{Model Abbreviations.} 
In the following tables, we use abbreviated model names for space efficiency: 
\textbf{o3} for OpenAI o3~\citep{openai_o3_2025}, 
\textbf{GPT-5.2} for GPT-5.2~\citep{openai_gpt52_2025}, 
\textbf{o4-mini} for o4-mini~\citep{openai_o4_mini_2025}, 
\textbf{Gem-3F} for Gemini-3-Flash~\citep{gemini3flash}, 
\textbf{Qwen3} for Qwen3-VL-235B-A22B-Thinking~\citep{qwen3vl}, and 
\textbf{IntVL3.5} for InternVL3.5-38B~\citep{chen2024internvl}.

\subsection{Performance by Scientific Domain}\label{app:perf-domain}

Table~\ref{tab:detailed-domain} shows model performance across 14 domain categories (13 individual arXiv primary categories plus an ``Other-Domain'' group covering seven low-frequency categories). We observe significant variation across domains. Proprietary models generally excel in domains such as q-bio and stat, while open-source models face the greatest challenges in Physics (physics) and Mathematics (math).

\begin{table}[h]
\centering
\footnotesize
\setlength{\tabcolsep}{3pt}
\caption{Model accuracy (\%) across scientific domains. Seven low-frequency domains (hep-th, q-fin, nucl-ex, math-ph, hep-lat, hep-ex, econ) are grouped as ``Other-Domain.''}
\label{tab:detailed-domain}
\begin{tabular}{lrrrrrr}
\toprule
\textbf{Domain} & \textbf{o3} & \textbf{GPT-5.2} & \textbf{o4-mini} & \textbf{Gem-3F} & \textbf{Qwen3} & \textbf{IntVL3.5} \\
\midrule
astro-ph & 73.00 & 52.00 & 61.00 & 76.00 & 68.00 & 56.00 \\
cond-mat & 85.00 & 58.00 & 68.00 & 81.00 & 64.00 & 54.00 \\
cs & 78.00 & 64.00 & 83.00 & 81.00 & 72.00 & 63.00 \\
eess & 80.00 & 56.00 & 83.00 & 82.00 & 70.00 & 56.00 \\
gr-qc & 66.67 & 63.00 & 79.00 & 83.00 & 66.00 & 55.00 \\
hep-ph & 80.95 & 60.00 & 73.00 & 83.00 & 62.00 & 52.00 \\
math & 72.00 & 62.00 & 79.00 & 81.00 & 77.00 & 49.00 \\
nlin & 61.29 & 58.00 & 78.00 & 83.00 & 71.00 & 55.00 \\
nucl-th & 77.78 & 53.00 & 78.00 & 82.00 & 63.00 & 50.00 \\
physics & 76.83 & 49.00 & 73.00 & 86.00 & 63.00 & 48.00 \\
q-bio & 87.18 & 48.00 & 74.00 & 89.00 & 68.00 & 53.00 \\
quant-ph & 90.24 & 55.00 & 80.00 & 75.00 & 61.00 & 60.00 \\
stat & 77.03 & 63.00 & 77.00 & 87.00 & 66.00 & 61.00 \\
Other-Domain & 70.67 & 55.00 & 79.00 & 85.00 & 61.00 & 54.00 \\
\midrule
\textbf{Overall} & \textbf{78.29} & \textbf{54.63} & \textbf{74.91} & \textbf{80.50} & \textbf{64.55} & \textbf{57.01} \\
\bottomrule
\end{tabular}
\end{table}

\subsection{Performance by Figure Type}\label{app:perf-figtype}

Table~\ref{tab:detailed-figtype} presents model performance across 9 figure type categories (three rare types---Illustration, Photo, and Pie Chart---are grouped as ``Other''). Graph and Bar Chart figures are generally easier for models, while Scatter Plot and Composite figures pose the greatest challenges.

\begin{table}[h]
\centering
\footnotesize
\setlength{\tabcolsep}{3pt}
\caption{Model accuracy (\%) across figure types, sorted by frequency in the dataset (most common first). Three rare types (Illustration, Photo, Pie Chart) are grouped as ``Other.''}
\label{tab:detailed-figtype}
\begin{tabular}{lrrrrrr}
\toprule
\textbf{Figure Type} & \textbf{o3} & \textbf{GPT-5.2} & \textbf{o4-mini} & \textbf{Gem-3F} & \textbf{Qwen3} & \textbf{IntVL3.5} \\
\midrule
Line Plot & 78.00 & 53.00 & 77.00 & 83.00 & 60.00 & 57.00 \\
Composite & 79.00 & 51.00 & 73.00 & 79.00 & 67.00 & 56.00 \\
Diagram & 75.00 & 62.00 & 77.00 & 79.00 & 71.00 & 57.00 \\
Scatter Plot & 78.57 & 48.00 & 63.00 & 71.00 & 63.00 & 58.00 \\
Bar Chart & 81.13 & 64.00 & 74.00 & 80.00 & 72.00 & 62.00 \\
Heatmap & 91.07 & 50.00 & 69.00 & 80.00 & 65.00 & 53.00 \\
Graph & 77.19 & 68.00 & 83.00 & 81.00 & 81.00 & 56.00 \\
Box Plot & 78.26 & 61.00 & 81.00 & 87.00 & 70.00 & 52.00 \\
Other & 73.24 & 62.00 & 77.00 & 84.00 & 70.00 & 60.00 \\
\midrule
\textbf{Overall} & \textbf{78.29} & \textbf{54.63} & \textbf{74.91} & \textbf{80.50} & \textbf{64.55} & \textbf{57.01} \\
\bottomrule
\end{tabular}
\end{table}

\subsection{Performance by Question Type}\label{app:perf-questype}

Table~\ref{tab:detailed-questype} shows model performance across five question types based on cognitive operations. Structural questions are generally easiest, while Compositional questions (multi-step reasoning) are most challenging. The largest proprietary--open-source gap appears for Comparative questions.

\begin{table}[h]
\centering
\footnotesize
\setlength{\tabcolsep}{3pt}
\caption{Model accuracy (\%) across 5 question types. Results reflect the cognitive complexity of different reasoning operations.}
\label{tab:detailed-questype}
\begin{tabular}{lrrrrrr}
\toprule
\textbf{Question Type} & \textbf{o3} & \textbf{GPT-5.2} & \textbf{o4-mini} & \textbf{Gem-3F} & \textbf{Qwen3} & \textbf{IntVL3.5} \\
\midrule
Relational & 76.00 & 51.00 & 80.00 & 82.00 & 54.00 & 57.00 \\
Comparative & 83.00 & 60.00 & 80.00 & 93.00 & 69.00 & 58.00 \\
Descriptive & 78.00 & 64.00 & 68.00 & 74.00 & 77.00 & 66.00 \\
Compositional & 71.00 & 51.00 & 63.00 & 73.00 & 66.00 & 48.00 \\
Structural & 86.36 & 70.00 & 84.00 & 88.00 & 79.00 & 64.00 \\
\midrule
\textbf{Overall} & \textbf{78.29} & \textbf{54.63} & \textbf{74.91} & \textbf{80.50} & \textbf{64.55} & \textbf{57.01} \\
\bottomrule
\end{tabular}
\end{table}

\section{Training Details}
\label{app:training-details}

This section provides additional training details not covered in \S~\ref{subsec:training} of the main paper.

\noindent\textbf{Base Model.}
We fine-tune \texttt{Qwen2.5-VL-7B-Instruct}~\citep{qwen2vl} using LoRA~\citep{hu2022lora} adapters (rank=16, $\alpha$=32) applied to all linear layers, while keeping the vision encoder and alignment module frozen.
We use a 9:1 train/validation split and select the best checkpoint based on validation performance.

\noindent\textbf{Training Configuration.}
The key hyperparameters are:
\begin{itemize}[leftmargin=*, itemsep=1pt]
    \item Training epochs: 1
    \item Learning rate: 4e-5 with cosine decay schedule
    \item Warmup ratio: 0.03
    \item Batch size: 3 per device with gradient accumulation steps of 5 (effective batch size: 60 across 4 GPUs)
    \item LoRA rank: 16, LoRA alpha: 32, target modules: all linear layers
    \item Optimizer: AdamW ($\beta_1=0.9$, $\beta_2=0.999$, weight decay=0.05)
    \item Mixed precision: bfloat16
    \item Maximum sequence length: 8{,}192 tokens
    \item Attention implementation: FlashAttention
    \item Gradient checkpointing: enabled
    \item Random seed: 42
\end{itemize}

\noindent\textbf{Computing Resources.}
All experiments were conducted on 4$\times$ NVIDIA H100 GPUs using the ms-swift framework~\citep{zhao2024swift}.

\section{Framework Instantiation with Data-Juicer}\label{app:data-pipeline}

To ensure easy reproducibility, we re-implement our framework instantiation (\S~\ref{subsec:instantiation}) using Data-Juicer~\citep{chen2024data,dj2}, a one-stop system to process multimodal data for foundation models.
We customize two new operators for Data-Juicer and orchestrate them with 7 existing built-in operators (detailed later).
This Data-Juicer implementation allows our work to be easily reproduced and scaled to larger data sizes.

\subsection{Custom Operator: LaTeX Figure Context Extractor}\label{app:custom-operator}

We customize \texttt{latex\_figure\_context\_extractor}, a reusable Data-Juicer operator that extracts figure-associated context \(\mathcal{P}\) from LaTeX documents.
Given a figure-caption pair \((F, C)\), the operator identifies and extracts paragraphs from LaTeX source that cite figure \(F\), providing context \(\mathcal{P}\) for claim extraction (Eq.~\ref{eq:claim-extraction}).
The operator can be applied to any LaTeX documents with figure references.

\noindent\textbf{Key Configuration Parameters:}
\begin{itemize}[leftmargin=*, noitemsep]
    \item \texttt{caption\_matching\_threshold}: Caption similarity threshold for figure matching (default: 0.9)
    \item \texttt{citation\_commands}: LaTeX reference commands to search (default: \verb|\ref|, \verb|\cref|, \verb|\autoref|)
    \item \texttt{latex\_cache\_path}: Path to LaTeX source folder
    \item \texttt{paragraph\_separator}: Pattern for delineating paragraphs (default: \verb|\n\n|)
\end{itemize}

This operator implements the same context extraction logic as described in Appendix~\ref{app:data-prep}, but in a reusable Data-Juicer operator format.

\subsection{Custom Operator: Claim Extractor}\label{app:custom-operator-claim}

We customize \texttt{claim\_extractor\_mapper}, a Data-Juicer operator that extracts atomic claims about a specific label from text paragraphs.
Given a label (e.g., \texttt{fig:results}, \texttt{tab:comparison}) and paragraph \(\mathcal{P}\) that references this label, this operator uses \(M_{\text{text}}\) to extract a set of atomic claims \(\mathcal{S} = \{s_j\}\) about the referenced object (Eq.~\ref{eq:claim-extraction}).
The operator can be applied to any labeled objects in text (figures, tables, equations, etc.).

\noindent\textbf{Key Configuration Parameters:}
\begin{itemize}[leftmargin=*, noitemsep]
    \item \texttt{section\_key}: Input field containing text paragraphs
    \item \texttt{label\_key}: Input field containing the target label
\end{itemize}

\subsection{Operator Orchestration}\label{app:pipeline-orchestration}

We re-implement the data preparation procedure detailed in Appendix~\ref{app:data-prep} and the Cross-Modal Verification framework (\S~\ref{subsec:instantiation}) using Data-Juicer operators.
Beyond our two custom operators (\S~\ref{app:custom-operator}, \S~\ref{app:custom-operator-claim}), all other steps leverage built-in Data-Juicer operators:

\noindent\textbf{Stage 1: Data Preparation.}
\begin{itemize}[leftmargin=*, noitemsep]
    \item \texttt{remove\_comments\_mapper}: Removes inline and multiline LaTeX comments
    \item \texttt{expand\_macro\_mapper}: Expands LaTeX macro definitions (e.g., \verb|\newcommand|, \verb|\def|)
    \item \texttt{remove\_bibliography\_mapper}: Removes bibliography sections to reduce noise
    \item \texttt{latex\_figure\_context\_extractor}: Extracts figure-associated context \(\mathcal{P}\) (\S~\ref{app:custom-operator})
\end{itemize}

\noindent\textbf{Stage 2: Generation.}
\begin{itemize}[leftmargin=*, noitemsep]
    \item \texttt{claim\_extractor\_mapper}: Extracts atomic claims \(\mathcal{S} = \{s_j\}\) from context \(\mathcal{P}\) using \(M_{\text{text}}\) (\S~\ref{app:custom-operator-claim}, implements Eq.~\ref{eq:claim-extraction})
    \item \texttt{generate\_qa\_from\_text\_mapper}: Generates questions and correct answers \((Q_j, A^*_j)\) from \(\mathcal{S}\) using \(M_{\text{text}}\) (implements Eq.~\ref{eq:qa-generation})
    \item \texttt{mllm\_mapper} (distractor generation): Generates plausible distractor options using \(M_{\text{vision}}^{\text{gen}}\) conditioned on \((Q_j, A^*_j, F, C)\), yielding the full candidate QA pair \(q_j = (Q_j, O_j, A^*_j)\) (implements Eq.~\ref{eq:distractor-generation})
\end{itemize}

\noindent\textbf{Stage 3: Verification.}
\begin{itemize}[leftmargin=*, noitemsep]
    \item \texttt{llm\_analysis\_filter} (source-consistency): Validates consistency with context $\mathcal{P}$ via $V_{\text{src}}$
    \item \texttt{llm\_analysis\_filter} (visual-dependence): Filters non-visual questions via $V_{\text{vis\_dep}}$
    \item \texttt{mllm\_mapper}: Generates answers using $M_{\text{vision}}^{\text{verify}}$ for each QA pair $q_j$
    \item \texttt{text\_pair\_similarity\_filter}: Compares model answers with $A^*_j$ via~$V_{\text{vis\_con}}$
\end{itemize}

\noindent\textbf{Vision-Based Verification Implementation.}
Since Data-Juicer does not provide a built-in LVLM filter, we implement $V_{\text{vis\_con}}$ through a two-step process: (1)~\texttt{mllm\_mapper} uses $M_{\text{vision}}^{\text{verify}}$ to generate selected options from $O_j$, and (2)~we filter by checking if the selected option matches~$A^*_j$.

\subsection{Code Release}\label{app:code-release}

We release the complete Data-Juicer configuration in YAML format and the implementations of our new operators.
These reusable operators are contributed to the official Data-Juicer repository, facilitating the community to reproduce, reuse, and extend the proposed pipeline.

\end{document}